\documentclass{article}

\usepackage[main, final]{neurips_2026}

\usepackage[utf8]{inputenc} 
\usepackage[T1]{fontenc}    
\usepackage{hyperref}       
\usepackage{url}            
\usepackage{booktabs}       
\usepackage{amsfonts}       
\usepackage{nicefrac}       
\usepackage{microtype}      
\usepackage{xcolor}         
\usepackage{multirow}
\usepackage{amsmath}
\usepackage{amsthm} 
\usepackage{graphicx}
\usepackage{listings}
\usepackage{subcaption}
\usepackage{wrapfig}
\usepackage{titletoc}
\usepackage[many]{tcolorbox}
\usepackage{xcolor}

\newtheorem{property}{Property}[section]

\definecolor{boxbg}{HTML}{F8F8F8}       
\definecolor{boxframe}{HTML}{3A3A3A}    
\definecolor{titlebg}{HTML}{D69A7A}     
\definecolor{poscolor}{HTML}{22A822}    
\definecolor{negcolor}{HTML}{D92A2A}    

\newtcolorbox{progressbox}[1][]{
    enhanced,                  
    colback=boxbg,             
    colframe=boxframe,         
    boxrule=1.5pt,             
    arc=4pt,                   
    top=15pt, bottom=10pt, left=10pt, right=10pt, 
    fonttitle=\large,          
    coltitle=white,            
    attach boxed title to top center={yshift=-\tcboxedtitleheight/2},
    boxed title style={
        colback=titlebg,       
        colframe=titlebg,      
        arc=3pt,               
        boxrule=0pt,
        top=4pt, bottom=4pt, left=15pt, right=15pt 
    },
    title={#1}                 
}
\newcommand{\posprog}[1]{\textcolor{poscolor}{#1}}
\newcommand{\negprog}[1]{\textcolor{negcolor}{#1}}

\usepackage[many]{tcolorbox} 
\usepackage{amsmath, amssymb} 
\usepackage{pifont} 
\usepackage{enumitem} 

\definecolor{mainborder}{RGB}{120, 120, 120} 
\definecolor{vanillared}{RGB}{139, 0, 0}     
\definecolor{stirgreen}{RGB}{0, 128, 0}      

\newtcolorbox{casebox}[1]{
    enhanced,
    colback=white,
    colframe=mainborder,
    boxrule=1pt,
    arc=2mm,
    top=4mm, bottom=4mm, left=4mm, right=4mm,
    title={#1},
    coltitle=black, 
    fonttitle=\large\bfseries,
    attach boxed title to top left={yshift=-\tcboxedtitleheight/2, xshift=5mm},
    boxed title style={
        colback=gray!20, 
        colframe=mainborder,
        boxrule=1.5pt,
        arc=1mm,
        left=2mm, right=2mm, top=1mm, bottom=1mm
    }
}

\newtcolorbox{vanillabox}[1]{
    enhanced,
    colback=white,
    colframe=vanillared,
    colbacktitle=vanillared,
    coltitle=white,
    title={\ding{55} \textbf{#1}},
    boxrule=1pt,
    arc=1mm,
    fonttitle=\small\sffamily,
    left=2mm, right=2mm, top=2mm, bottom=2mm,
    width=0.485\linewidth,         
    nobeforeafter,                 
    equal height group=mycasebox   
}

\newtcolorbox{stirbox}[1]{
    enhanced,
    colback=white,
    colframe=stirgreen,
    colbacktitle=stirgreen,
    coltitle=white,
    title={\ding{51} \textbf{#1}},
    boxrule=1pt,
    arc=1mm,
    fonttitle=\small\sffamily,
    left=2mm, right=2mm, top=2mm, bottom=2mm,
    width=0.485\linewidth,         
    nobeforeafter,                 
    equal height group=mycasebox   
}

\title{Efficient Reasoning via Constrained Optimization in Latent Space}

\author{%
  Zhinan Hou $^1$ ~~~
  Xingchen Li$^1$ ~~~
  Keyou You$^1$\thanks{Corresponding author: youky@tsinghua.edu.cn} \\\\
  $^1$Tsinghua University
}

\begin{document}

\maketitle

\begin{abstract}
Large Reasoning Models (LRMs) have shown remarkable reasoning capabilities, yet they still suffer from overthinking, generating redundant reasoning steps which incur substantial token consumption. Existing methods, such as suppressing reflective keywords or forcing shorter reasoning lengths, attempt to mitigate this issue but inevitably truncate necessary steps and induce underthinking, thereby compromising performance. To address this dilemma, we investigate the latent representations and observe that efficient reasoning steps naturally cluster into a concentrated region in latent space, while those deviating from this region tend to produce verbose sequences. To leverage this, we keep reasoning focused within this region via a quadratic program which projects deviating hidden states back into the region. Then we propose a novel training-free framework to achieve efficient reasoning that reduces token generation costs without sacrificing performance. Extensive experiments conducted on four models ranging from 1.5B to 14B, and across six benchmarks in math reasoning, coding, and scientific QA, validate the effectiveness of our method, up to a 12.1\% improvement in accuracy while reducing generated tokens by 11.8\% to 52.8\%. Codes are available at \href{https://github.com/hzn18/Opt4Reasoning}{https://github.com/hzn18/Opt4Reasoning}.
\end{abstract}

\section{Introduction}
Large Reasoning Models (LRMs) have significantly enhanced the ability of artificial intelligence to solve complex tasks \citep{novikov2025alphaevolve, guo2025deepseek}. A key technique for this success is Chain of Thought (CoT) reasoning, which enables models to dynamically scale test time computation by generating a sequence of intermediate steps before producing the final answer \citep{wei2022chain}. This allows the model to break down complex problems into more manageable components, leading to more reliable and accurate outcomes.

However, long CoT introduces significant inference overhead due to the autoregressive nature of these models. Moreover, excessively long reasoning processes frequently induce \textit{overthinking}, a phenomenon where models become trapped in problematic behaviors like meaningless self-hesitation, infinite loops, and redundant verbosity, which finally compromise accuracy \citep{wu2025more}. To address this issue, recent studies have attempted to shorten reasoning chains through techniques such as early exit \citep{wu2025more, yang2025dynamic} or compression via reinforcement learning and supervised fine-tuning \citep{huang2025adactrl, li2026making}. Yet, these approaches tend to bias models toward pursuing overly short reasoning paths. As a result, they may inevitably abandon valuable reasoning steps and degrade overall performance, inducing the problem of \textit{underthinking} \citep{liefficient}. Therefore, a central challenge is \textit{how to achieve efficient reasoning that reduces token consumption while maintaining or even improving reasoning performance.}

To achieve this, it is crucial to understand how efficient reasoning differs from redundant generation at the representation level. Accordingly, we analyze the hidden representations of reasoning steps in the latent space. We discover that efficient reasoning steps naturally form a concentrated region while redundant steps such as meaningless double-checking or irrelevant computations, usually scatters around this region. Then, we examined the geometric distance of the entire reasoning trajectory to this concentrated region. Our analysis reveals that the further a trajectory deviates from this region, the higher the probability that the model produces an incorrect answer or generates an exceptionally long sequence of tokens.

Inspired by these geometric insights, we aim to keep reasoning focused within the concentrated region of the latent space. During the inference process, if the hidden state of the model attempts to deviate from this region, we correct it by projecting the state to the closest point within the region, which is formulated as a Quadratic Programming (QP) problem. This allows us to steer the reasoning back into the target region while preserving the essential semantic information of the current task. In doing so, the model is implicitly guided to generate more effective reasoning steps. Interestingly, this methodology naturally aligns with biological mechanisms of advanced cognition, which similarly depend on the \textit{inhibition} of neural signals to prevent cognitive activity from descending into unproductive divergence \citep{yizhar2011neocortical}.

We conducted extensive experiments across four models ranging from 1.5B to 14B to verify the effectiveness of our method. First, we tested it on mathematical datasets with different levels of difficulty including GSM8K, MATH 500, AMC 2023, and AIME 2025. The results show that our method achieves a token reduction while consistently improving model accuracy. Moreover, it also demonstrates strong cross-domain generalization capability in tasks such as LiveCodeBench (code generation) and Diamond GPQA (knowledge-based QA). Furthermore, we integrate the method into the vLLM engine and achieve a 24.9\% to 38.4\% reduction in end-to-end inference latency, highlighting its high efficiency for real-world deployment. 

\section{Related Work}
Efforts to achieve efficient reasoning in LRMs have gained significant traction to enhance both inference efficiency and output quality. Among them, training-based methods typically use supervised fine-tuning to compress the reasoning process \cite{yuan2025not, yu2025long} or use reinforcement learning to enable the model to adaptively generate chains of thought of different lengths \cite{luo2025autol2s, jiang2025think}. However, these methods incur significant computational costs and are orthogonal to inference-time interventions like ours, so we do not directly compare them in this work. Existing training-free methods
include early exiting \cite{yang2025dynamic, zhang2025reasoning, qiao2025concise} and representation steering \cite{liefficient, chenseal, shi2026internalizing, huang2025mitigating, nguyen2026atlas}. Specifically, representation steering dynamically adjusts the reasoning trajectory by directly intervening in the model's latent activation space during inference. Our work falls into this category. Existing representation steering methods are usually reactive and merely push the latent state away from problematic behaviors (e.g. overthinking), which inevitably risks over-correction and induces underthinking. Differently, our work explicitly considers \textit{where the efficient reasoning state should actually reside}. We reveal the geometric concentration of efficient reasoning within the latent space and proactively project deviant hidden states back into this concentrated region via constrained optimization. More related works can be seen in Appendix \ref{app:related_work}.

\section{Problem Formulation and Our Objective} 
Given a question $q$, a large reasoning model, denoted as $\pi$, autoregressively generates a sequence of intermediate reasoning steps $c = [c_1, c_2, \dots, c_N]$ before producing the final answer $y$. This process can be formally expressed as:
\begin{equation}
    c_t \sim \pi(\cdot | c_{<t}, q), \quad y \sim \pi(\cdot | c, q)
\end{equation}
where $c_t$ is the reasoning step generated at time step $t$, and $c_{<t}$ represents all preceding steps. 

Naturally, we are interested in the contribution of one individual reasoning step to the final outcome. To quantify this contribution, we evaluate its impact on the model's expected correctness and remaining generation length. Specifically, after observing a partial trajectory up to step $t$ (denoted as $c_{\le t}$), the expected probability of the model ultimately reaching the correct answer $y^*$ can be written as $\mathbb{P}_{(c_{>t}, y) \sim \pi(\cdot | c_{\le t}, q)}(y = y^*)$. Similarly, the expected number of remaining tokens required to complete the generation from this step is given by $\mathbb{E}_{(c_{>t}, y) \sim \pi(\cdot | c_{\le t}, q)}\text{Len}(c_{>t}, y)$.

By comparing these expected values before step $c_t$ (conditioned on $c_{<t}$) and after step $c_t$ (conditioned on $c_{\le t}$), we can isolate the  contribution of $c_t$. Thus, we introduce the concept of \textbf{progress}, a two-dimensional vector $p(c_t; c_{<t}, q) = [p_\text{acc}, p_\text{len}]$, which explicitly quantifies the impact of step $c_t$ on both the final accuracy and the generation length:
\begin{equation} \label{eq:progress}
    \begin{aligned}
    p_\text{acc} &= \mathbb{P}_{(c_{>t}, y) \sim \pi(\cdot | c_{\le t}, q)}(y = y^*) - \mathbb{P}_{(c_{\ge t}, y) \sim \pi(\cdot | c_{<t}, q)}(y = y^*) \\
    p_\text{len} &= \mathbb{E}_{(c_{\ge t}, y) \sim \pi(\cdot | c_{<t}, q)}\text{Len}(c_{\ge t}, y) - \mathbb{E}_{(c_{>t}, y) \sim \pi(\cdot | c_{\le t}, q)}\text{Len}(c_{>t}, y) - \text{Len}(c_t)
    \end{aligned}
\end{equation}
where $y^*$ is the ground-truth answer, and $\mathrm{Len}(\cdot)$ denotes the token length. 

In Eq. \eqref{eq:progress}, $p_\text{acc}$ represents the change in the probability of reaching the correct answer $y^*$, where a positive value indicates the step has moved the model closer to success. $p_\text{len}$ measures the reduction in the total expected token length. A positive value implies that the step has shortened the overall path to the solution. This formulation naturally captures typical reasoning behaviors. For instance, a negative $p_\text{acc}$ reflects a drop in correctness, which is often seen in cases where the reasoning is too shallow or terminates prematurely, i.e., \textit{underthinking}. Similarly, a negative $p_\text{len}$ means that the token cost of the step is greater than the amount it reduces from the remaining reasoning process. Such steps usually correspond to the behaviors in \textit{overthinking} such as unnecessary computations or redundant double-checking, which inevitably increase the overall length of the reasoning process.

\textbf{Empirical Example for Understanding Progress.} To further illustrate the concept of progress, the following example presents a reasoning trajectory generated by DeepSeek-R1-Distill-Qwen-1.5B. For each reasoning step $c_t$, we sample the model 10 times conditioned on $c_{\le t}$ and $c_{<t}$, then compute the differences in the accuracy and the expected remaining length as the estimation of progress. We find that steps with negative progress are often associated with self-correction loops, hesitation, or repetitive reformulation, which increase token cost without improving correctness. In contrast, steps with positive progress typically correspond to necessary computation or preliminary conclusions that move the solution forward.

\begin{progressbox}[Example 1: Empirical Example of Progress in CoT]

\textsc{Question:}\\[4pt]
How many integers $n$ satisfy the condition $100 < n < 200$ and the condition $n$ has the same remainder whether it is divided by 6 or by 8?
\vspace{12pt}

\textsc{Progess in CoT:}\\[4pt]
\textit{\posprog{($p$ = [0.0, +1322])} Okay, so I have this problem here: I need to find how many integers $n$ satisfy two conditions. ...} \textit{\negprog{($p$ = [-0.1, -612])} Alright, so let's break it down. The first condition is straightforward: ... So, mathematically, this can be written as: $n \equiv r \pmod 6$ and $n \equiv r \pmod 8$ ...} \textit{\negprog{($p$ = [0.0, -620])} But wait, hold on. Let me check: if r is the remainder when n is divided by 6, then it's 0 $\le$ r < 6. ...} \textit{\posprog{($p$ = [+0.1, +539])}  So, for each r, we can find the number of k such that 24k + r is between 101 and 199.} \textit{\negprog{($p$ = [0.0, -941])} So, let's solve for k in each case.\textbackslash n\textbackslash n ...} \textit{\posprog{($p$ = [+0.1, +818])} But n=24*4 +5=96 +5=101, which is the lower bound. So, k=4,5,6,7,8, which is 5 values. ...} \textit{\negprog{($p$ = [-0.1, -827])} Wait, 6 is not equal to 0. So, that's a problem. Wait, so 102 is not included. ...} \textit{\negprog{($p$ = [-0.0, -669])} Alternatively, perhaps we can think in terms of the least common multiple.}\textit{\posprog{($p$ = [0.0, +15])} **Final Answer**The number of integers n that satisfy the given conditions is \textbackslash \textbackslash boxed\{25\}.}

\end{progressbox}

Intuitively, an efficient reasoning step $c_t$ should simultaneously improve accuracy and promote generation efficiency. We formalize this by defining the set $S(c_{<t}, q)$ as:
\begin{equation} \label{obj:set}
    S(c_{<t}, q) := \{c_t \mid p(c_t; c_{<t}, y) \ge [0,0]\}
\end{equation}
Each step in the set $S(c_{<t}, q)$ is both beneficial for reaching the correct answer and efficient in terms of generation length. Consequently, our objective is to guide the LRM to favor steps within the set \eqref{obj:set}, thereby ensuring that the resultant reasoning trajectory is both accurate and concise.

\section{Analysis of Reasoning Steps in Latent Space} \label{sec:analysis}
The definition in Eq.~\eqref{obj:set} provides a conceptual formulation for efficient reasoning. However, explicitly constructing this set is computationally prohibitive. Evaluating the progress vector $p$ requires extensive rollouts, incurring massive computational overhead. As a practical alternative, we seek an empirical proxy for this target set.

\subsection{A Geometric Perspective on Efficient Reasoning}
A natural way is to leverage LRM itself to generate efficient reasoning paths. Specifically, we prompt the model with the ground-truth answer and conciseness instructions (see Appendix \ref{app:direct}), encouraging it to produce compact deductions as final outputs. Then we discard the intermediate \texttt{<think>} block and retain only the finalized output after \texttt{</think>}. As observed in the specific example provided in Appendix \ref{app:pro_example}, the steps in the output typically exhibit positive progress and closely align with the definition of our target set $S(c_{<t}, q)$. We define such outputs as \textit{Direct CoT}. For comparison, we also collect responses generated from the question alone, i.e., \textit{Vanilla CoT}.
 
We conduct experiments on the MATH dataset \citep{hendrycks2measuring} using DeepSeek-R1-Distill-Qwen-1.5B \citep{guo2025deepseek}. Specifically, we randomly sample 300 questions from the default training set, generating one Direct CoT and one Vanilla CoT for each question. During the inference process, we concatenate these CoTs with the prompt of Vanilla CoT and then extract the hidden representations corresponding to the step delimiters (e.g., ``\textbackslash n \textbackslash n'') across different model layers. We then apply Principal Component Analysis (PCA) to project these hidden representations into a low-dimensional latent space. 

\begin{figure}[ht]
    \centering
    \includegraphics[width=\textwidth]{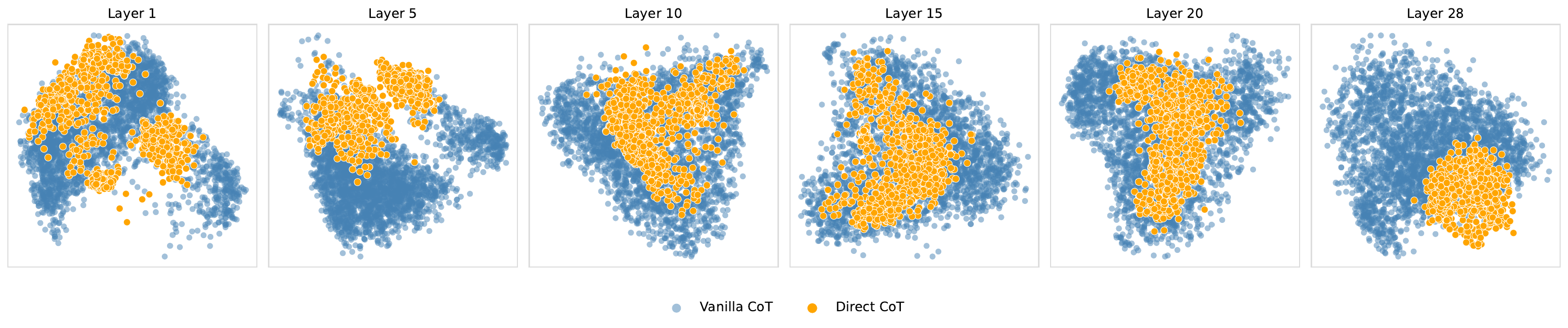} 
    \caption{Visualization of different reasoning steps in the latent space. The horizontal and vertical axes correspond to the first and second PCA components, respectively.}
    \label{fig:pca}
\end{figure}

As shown in Figure \ref{fig:pca}, the representations of Direct CoTs form a concentrated region, whereas those of Vanilla CoTs are dispersed and encapsulate this inner region. Moreover, as shown in Figure \ref{fig:further_visual},  the Vanilla CoT representations falling outside this concentrated region likely correspond to redundant steps such as getting stuck in repetitive loops or redundant self-verification. Motivated by this geometric observation, we adopt the concentrated region formed by Direct CoTs as our empirical proxy for the target set in Eq.~\eqref{obj:set}. For clarity, we refer to this region as the \textbf{efficient reasoning set}, denoted by $\mathcal{H}$. We also include additional visualizations for other models in Appendix \ref{app:other_visual}.

\subsection{Computation of the Efficient Reasoning Set $\mathcal{H}$} \label{subsec:proxy}

We now detail the computation method for the set $\mathcal{H}$. At a high level, we project the high-dimensional hidden representations into a lower-dimensional latent space, and subsequently compute boundary hyperplanes to enclose the target set $\mathcal{H}$. Specifically, let $\mathbf{X}_{D}$ and $\mathbf{X}_{V}$ denote the hidden representations of Direct and Vanilla CoTs, respectively. We first project these into a $k$-dimensional latent space via a PCA matrix $\mathbf{U} \in \mathbb{R}^{k \times d}$ ($k \ll d$), yielding $\mathbf{Z}_{D} = \mathbf{U}\mathbf{X}_{D}$ and $\mathbf{Z}_{V} = \mathbf{U}\mathbf{X}_{V}$. The intuition for constructing $\mathcal{H}$ is to maximize the inclusion of $\mathbf{Z}_{D}$ while maximally excluding $\mathbf{Z}_{V}$. Because the dispersed $\mathbf{Z}_{V}$ representations geometrically encapsulate the concentrated $\mathbf{Z}_{D}$ region, we aim to establish hyperplanes from multiple directions to enclose this target region.

To find these directions, we first apply K-Means to partition the $\mathbf{Z}_{V}$ representations into $M$ distinct clusters $\{\mathbf{Z}_{V}^{(1)}, \dots, \mathbf{Z}_{V}^{(M)}\}$. For each cluster $m \in \{1, \dots, M\}$, we train a linear SVM to separate the target region $\mathbf{Z}_{D}$ from this specific cluster $\mathbf{Z}_{V}^{(m)}$. This procedure yields $M$ initial half-spaces, each defined by a directional boundary $\mathbf{w}_m^\top \mathbf{z} + b_m \le 0$. We then further relax each hyperplane outward by adjusting its intercept as follows:
\begin{equation} \label{eq:relax}
    b'_m = - \max \left( P_{\alpha}\left(\{ \mathbf{w}_m^\top \mathbf{z} \mid \mathbf{z} \in \mathbf{Z}_{D} \}\right), \ \mathbf{w}_m^\top \bar{\mathbf{z}}_{D} \right)
\end{equation}
where $P_{\alpha}(\cdot)$ denotes the $\alpha$-th percentile of the projected scalar values, and $\bar{\mathbf{z}}_{D}$ is the mean vector of the $\mathbf{Z}_{D}$. These two components serve different purposes. The first term allows the shifted boundary to retain at least an $\alpha$ proportion of the Direct CoT steps. The second term ensures that the boundary never excludes the center of the Direct CoT. 

Finally, the efficient reasoning set $\mathcal{H}$ can be obtained through the intersection of the half-spaces bounded by these $M$ adjusted hyperplanes:
\begin{equation}
    \mathcal{H} = \left\{ \mathbf{z} \in \mathbb{R}^k \mid \mathbf{w}_m^\top \mathbf{z} + b'_m \le 0, \quad \forall m \in \{1, \dots, M\} \right\}
\end{equation}
This formulation bounds the Direct CoT reasoning space into a convex polytope. Specially, because the mean vector $\bar{\mathbf{z}}_{D}$ satisfies all $M$ linear constraints by design, it serves as a guaranteed feasible point, ensuring that the intersection $\mathcal{H}$ is strictly non-empty.

\subsection{Reasoning Performance and Distance to $\mathcal{H}$}

We further investigate what happens when steps in a reasoning sequence frequently fall outside the region $\mathcal{H}$. Intuitively, if $\mathcal{H}$ represents the efficient reasoning, then deviating from $\mathcal{H}$ could correspond to inefficient reasoning, such as generating redundant steps. Let a generated trajectory be denoted as $c=[c_1,\dots,c_N]$, with $z_i$ representing the latent representation of the step $c_i$. We define the distance from the reasoning trajectory to the set $\mathcal{H}$ as:
\begin{equation}
    d(c,\mathcal{H})=\frac{1}{N}\sum_{i=1}^{N}\min_{u\in\mathcal{H}}\lVert z_i-u\rVert_2
\end{equation}

\begin{figure}[htbp]
    \centering
    \begin{subfigure}[b]{0.48\textwidth}
        \centering
        \includegraphics[width=\textwidth]{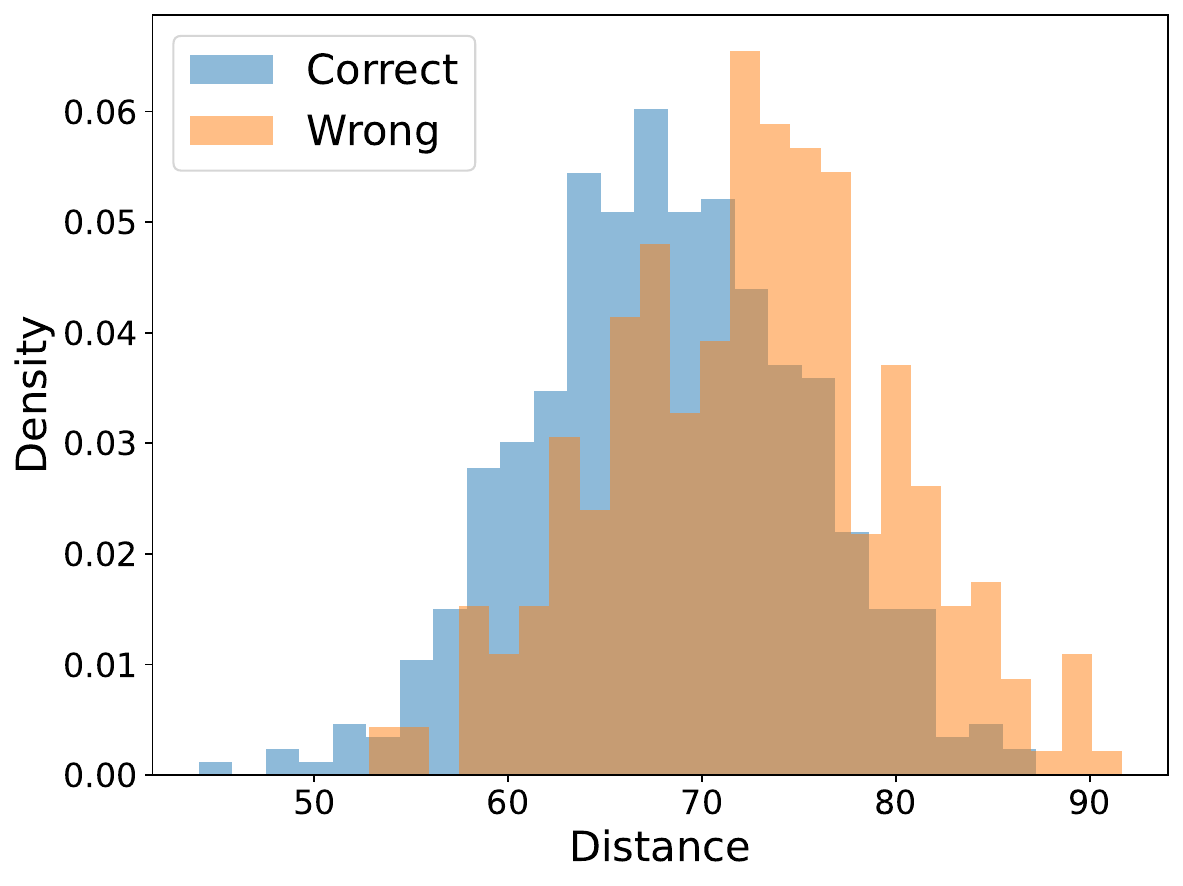} 
        \caption{Reasoning Accuracy vs. Distance}
        \label{fig:sub_a}
    \end{subfigure}
    \hfill 
    \begin{subfigure}[b]{0.48\textwidth}
        \centering
        \includegraphics[width=\textwidth]{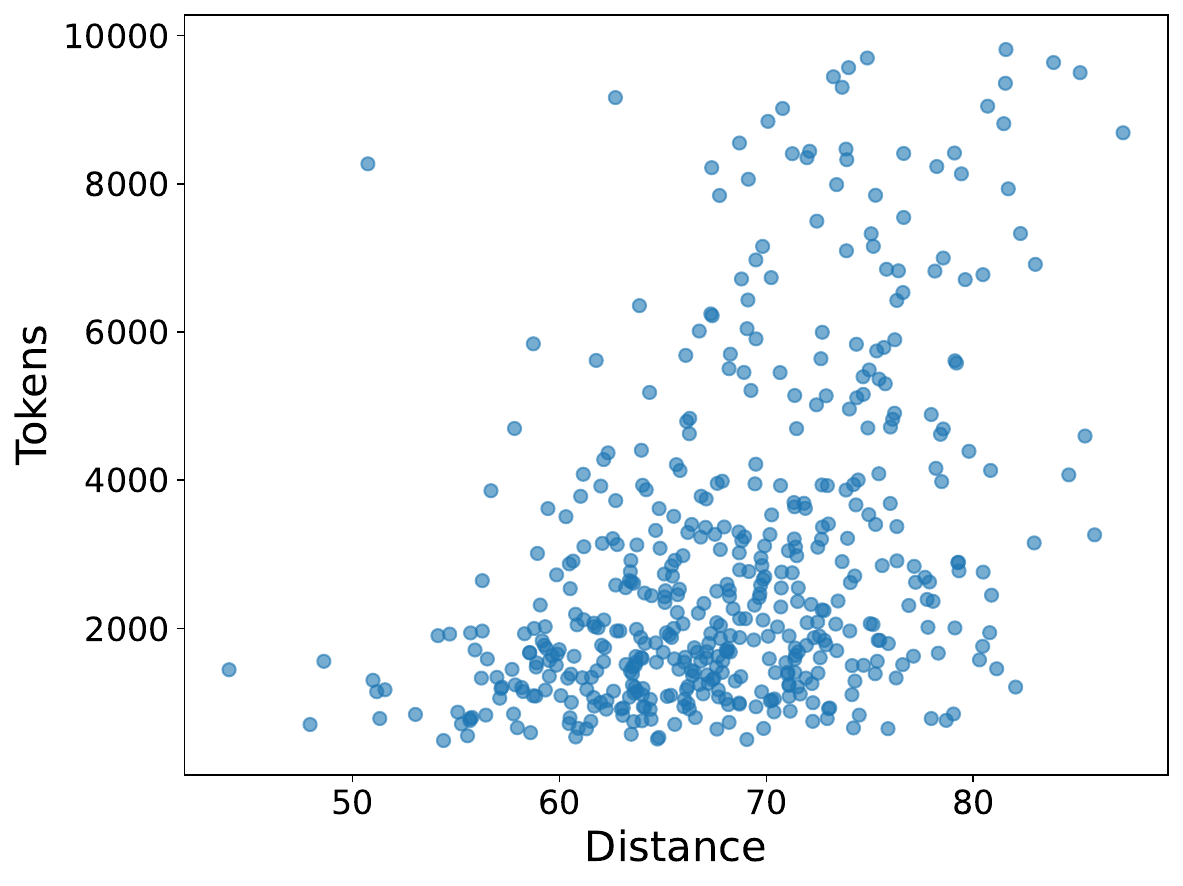} 
        \caption{Total Token Length vs. Distance}
        \label{fig:sub_b}
    \end{subfigure}
    
    \caption{(a) The distribution of distances to the set $\mathcal{H}$ for trajectories with correct and incorrect answers. (b) The relation between the distance $d(c,\mathcal{H})$ and the total token length of the reasoning trajectory.}
    \label{fig:obs}
\end{figure}

Then, we compute the distance $d(c,\mathcal{H})$ of each previously collected \textit{Vanilla CoT} trajectory and evaluate how it relates to reasoning performance. As illustrated in Figure~\ref{fig:obs}, greater distance indicates worse reasoning outputs. Specifically, Figure~\ref{fig:obs}(a) shows that correct reasoning trajectory generally stays close to the set $\mathcal{H}$, whereas incorrect ones drift much further away. Additionally, Figure~\ref{fig:obs}(b) reveals a clear positive correlation (Spearman's $\rho = 0.4977$) between $d(c,\mathcal{H})$ and total token length, indicating that sequences further from the set $\mathcal{H}$ may be unnecessarily long. Collectively, these findings indicate that deviating from the set  $\mathcal{H}$ corresponds to a higher risk of failure and computational overthinking.

\section{Representation Steering via Constrained Optimization} \label{sec:method}
Building upon the observation above, we propose a novel steering mechanism. Our goal is to actively correct the model's reasoning trajectory by steering its hidden states back into the efficient reasoning region bounded by $\mathcal{H}$. 

Specifically, we apply the intervention exclusively to the step delimiter tokens ``\textbackslash n \textbackslash n''.  Let $\mathbf{x}_t \in \mathbb{R}^d$ denote its hidden representation of such a token at a specific layer. We first project this representation into our $k$-dimensional latent space via PCA, obtaining $\mathbf{z}_t = \mathbf{U}\mathbf{x}_t$. If $\mathbf{z}_t$ falls outside the set $\mathcal{H}$, we seek the closest point within $\mathcal{H}$ to serve as the corrected latent state. Since $\mathcal{H}$ is a convex polytope defined by the intersection of $M$ half-spaces, finding this closest point $\mathbf{z}^*_t$ naturally formulates as a constrained Quadratic Programming (QP) problem:
\begin{equation} \label{eq:projection}
    \mathbf{z}^*_t = \arg\min_{\mathbf{u} \in \mathbb{R}^k} \frac{1}{2} \lVert \mathbf{z}_t - \mathbf{u} \rVert_2^2 \quad \text{s.t.} \quad \mathbf{w}_m^\top \mathbf{u} + b'_m \le 0, \quad \forall m \in \{1, \dots, M\}
\end{equation}
For simplicity, we denote this projection operation as $\mathbf{z}^*_t = \text{Proj}_{\mathcal{H}}(\mathbf{z}_t)$. As established in Section \ref{subsec:proxy}, the set $\mathcal{H}$ is guaranteed to be non-empty (containing at least the mean vector $\bar{\mathbf{z}}_{dir}$). Consequently, this QP problem is strictly feasible and admits a unique global optimum.

Once the corrected latent representation is computed, we map this correction back to the original high-dimensional space. Notably, simply replacing the original representation with the inverse PCA transform (i.e., $\mathbf{U}^\top\text{Proj}_{\mathcal{H}}(\mathbf{U}\mathbf{x}_t)$) would discard all information residing in the orthogonal complement of the PCA subspace, potentially destroying task-specific semantic details. Therefore, we preserve the orthogonal residual and only modifies the components within the principal subspace as follow:
\begin{equation} 
    \mathbf{x}_t \leftarrow \mathbf{x}_t + \lambda \mathbf{U}^\top \big( \text{Proj}_{\mathcal{H}}(\mathbf{U}\mathbf{x}_t) - \mathbf{U}\mathbf{x}_t \big)
\end{equation}
where $\lambda$ is a positive coefficient to control the intervention strength. Notably, we introduce $\lambda$ into the formulation primarily to facilitate sensitivity analysis (see Section \ref{sec:ablation}) rather than as a hyperparameter requiring tuning. We fix $\lambda = 1.0$ as the default setting.

Furthermore, the computational overhead of this intervention is marginal compared to the original forward pass. This stems from two factors. First, the intervention is infrequent, targeting only the few step delimiters within a sequence; Second, the QP projection is computationally efficient by construction, as it operates in a low-dimensional latent space with a limited number of constraints. In our implementation, the problem involves only 10–16 decision variables and 8 linear constraints. 

\textbf{GPU Acceleration for Solving QP.} 
Despite its low mathematical complexity, QP solvers typically execute on the CPU, which introduces additional latency due to the frequent data transfer between the CPU and GPU. This could noticeably degrade the overall throughput, especially in a highly optimized LRM inference pipeline. To eliminate this bottleneck, we accelerate the optimization process by exploiting the structural properties of Multi-Parametric Quadratic Programming (mp-QP). Specifically, mp-QP theory establishes that the optimal solution $\mathbf{z}^*_t$ can be analytically expressed as a continuous piecewise linear function of the input parameter $\mathbf{z}_t$ \citep{bemporad2002explicit}. Practically, this allows us to partition the continuous latent space into a finite set of predefined regions offline. During inference, solving the specific QP instance is reduced to querying the region index and applying its corresponding affine transformation matrix. We precompute and deploy these mappings directly on the GPU as tensor operations, eliminating the CPU-GPU communication overhead and ensuring high-throughput reasoning. A detailed introduction of the quadratic programming problem and the explicit acceleration via offline mp-QP are provided in Appendix \ref{app:details_on_qp}.

\section{Experiments} \label{sec:exp}
\begin{table}[ht!]
\centering
\caption{Performance comparison against baselines. For each column within a model group, the best result is \textbf{bolded} and the second-best is \underline{underlined}.}
\label{tab:combined_performance}
\small
\renewcommand{\arraystretch}{1.2}
\setlength{\tabcolsep}{2.5pt}
\resizebox{\textwidth}{!}{
\begin{tabular}{lcccccccccccc}
\toprule
\multirow{3}{*}{\textbf{Methods}} & \multicolumn{8}{c}{\textbf{MATH}} & \multicolumn{2}{c}{\textbf{PROGRAMMING}} & \multicolumn{2}{c}{\textbf{SCIENCE}} \\
\cmidrule(lr){2-9} \cmidrule(lr){10-11} \cmidrule(lr){12-13}
 & \multicolumn{2}{c}{GSM8K} & \multicolumn{2}{c}{MATH500} & \multicolumn{2}{c}{AMC2023} & \multicolumn{2}{c}{AIME2025} & \multicolumn{2}{c}{LiveCodeBench} & \multicolumn{2}{c}{GPQA-Diamond} \\
\cmidrule(lr){2-3} \cmidrule(lr){4-5} \cmidrule(lr){6-7} \cmidrule(lr){8-9} \cmidrule(lr){10-11} \cmidrule(lr){12-13}
 & Acc. & Tok. & Acc. & Tok. & Acc. & Tok. & Acc. & Tok. & Acc. & Tok. & Acc. & Tok. \\
\midrule
\multicolumn{13}{l}{\textbf{DeepSeek-R1-Distill-Qwen-1.5B}} \\
Vanilla & 81.5 & 1489 & 82.1 & 4476 & 64.4 & 7708 & 17.3 & 11811 & \underline{32.1} & 10013 & 33.3 & 8241 \\
CCoT & 76.6 & \textbf{577} & 82.0 & 3832 & 68.8 & 7012 & 22.7 & 11077 & 30.6 & 9824 & 33.3 & 7592 \\
DEER & 74.2 & \underline{672} & 72.2 & \textbf{2578} & 62.5 & \underline{5139} & 20.2 & \underline{9601} & 23.7 & 8227 & 29.7 & 7655 \\
SEAL & \underline{82.2} & 924 & 82.4 & 3460 & \underline{72.5} & 5382 & 23.5 & 9904 & 30.5 & \underline{8111} & \underline{33.8} & \underline{6620} \\
STIR & 81.2 & 909 & 79.8 & 3363 & 70.0 & 6157 & 23.1 & 10572 & 31.7 & 9632 & 26.8 & 7027 \\
ReBalance & 81.0 & 895 & \underline{83.0} & 3697 & 67.5 & 6292 & \underline{24.4} & 10600 & 30.2 & 9196 & 22.7 & 9071 \\
Ours & \textbf{83.1} & 853 & \textbf{83.2} & \underline{2949} & \textbf{75.2} & \textbf{4821} & \textbf{26.7} & \textbf{8502} & \textbf{32.7} & \textbf{7656} & \textbf{36.4} & \textbf{5651} \\
$\Delta$ vs. Van. & \textcolor{orange}{(+1.6)} & \textcolor{blue}{(-42.7\%)} & \textcolor{orange}{(+1.1)} & \textcolor{blue}{(-34.1\%)} & \textcolor{orange}{(+10.8)} & \textcolor{blue}{(-37.5\%)} & \textcolor{orange}{(+9.4)} & \textcolor{blue}{(-28.0\%)} & \textcolor{orange}{(+0.6)} & \textcolor{blue}{(-23.5\%)} & \textcolor{orange}{(+3.1)} & \textcolor{blue}{(-31.4\%)} \\
\midrule
\multicolumn{13}{l}{\textbf{DeepSeek-R1-Distill-Qwen-7B}} \\
Vanilla & 90.9 & 1129 & 91.6 & 3647 & 86.3 & 5846 & 26.9 & 11209 & 58.9 & 8320 & 48.9 & 7194 \\
CCoT & 89.4 & \textbf{679} & 91.8 & 3225 & 90.2 & 5522 & 33.8 & 10695 & 59.5 & 8077 & 50.5 & 7101 \\
DEER & 90.1 & 753 & 89.6 & \textbf{2245} & 85.0 & \underline{4107} & 31.9 & 9442 & 54.2 & 7867 & 49.1 & 7166 \\
SEAL & \underline{91.0} & 825 & 90.6 & 2838 & 85.8 & 4813 & 32.1 & \underline{9422} & 57.1 & \underline{7060} & 50.5 & \underline{5456} \\
STIR & 90.6 & 730 & 88.2 & 2669 & 85.0 & 5244 & \underline{36.5} & 10303 & \underline{59.7} & 7739 & 46.9 & 6035 \\
ReBalance & 90.5 & 1016 & \underline{92.1} & 3384 & \textbf{92.5} & 5075 & 33.5 & 10258 & 55.0 & 7749 & \textbf{52.0} & 6833 \\
Ours & \textbf{91.3} & \underline{716} & \textbf{92.2} & \underline{2599} & \underline{91.1} & \textbf{4083} & \textbf{39.0} & \textbf{9042} & \textbf{61.2} & \textbf{6788} & \underline{51.0} & \textbf{4599} \\
$\Delta$ vs. Van. & \textcolor{orange}{(+0.4)} & \textcolor{blue}{(-36.6\%)} & \textcolor{orange}{(+0.6)} & \textcolor{blue}{(-28.7\%)} & \textcolor{orange}{(+4.8)} & \textcolor{blue}{(-30.2\%)} & \textcolor{orange}{(+12.1)} & \textcolor{blue}{(-19.3\%)} & \textcolor{orange}{(+2.3)} & \textcolor{blue}{(-18.4\%)} & \textcolor{orange}{(+2.1)} & \textcolor{blue}{(-36.1\%)} \\
\midrule
\multicolumn{13}{l}{\textbf{Qwen3-4B}} \\
Vanilla & 94.9 & 2476 & 95.2 & 4587 & 95.9 & 7588 & 60.2 & 16835 & \underline{87.7} & 7818 & 53.2 & 8080 \\
CCoT & 94.8 & \underline{1237} & \underline{95.4} & \underline{3834} & 96.6 & 6518 & 63.8 & 16034 & 86.7 & 7160 & 53.5 & 6836 \\
DEER & 94.3 & 1331 & 94.8 & 3947 & 95.0 & \underline{6204} & 62.1 & 15972 & 76.2 & 6993 & 53.0 & 7744 \\
SEAL & 95.0 & 1275 & 94.6 & 4015 & 90.8 & 6667 & \underline{64.4} & 14944 & 87.0 & \textbf{6559} & 50.0 & \underline{6739} \\
STIR & 94.4 & 1387 & 91.4 & 4075 & 92.5 & 6798 & 57.9 & \textbf{12768} & 84.0 & 7170 & \underline{55.0} & 6792 \\
ReBalance & \underline{95.3} & 1645 & 95.2 & 4137 & \underline{97.0} & 6289 & 60.6 & 15008 & 85.5 & 7475 & 53.9 & 7038 \\
Ours & \textbf{95.4} & \textbf{1168} & \textbf{95.8} & \textbf{3550} & \textbf{97.5} & \textbf{5898} & \textbf{67.1} & \underline{14809} & \textbf{88.5} & \underline{6898} & \textbf{57.5} & \textbf{6698} \\
$\Delta$ vs. Van. & \textcolor{orange}{(+0.5)} & \textcolor{blue}{(-52.8\%)} & \textcolor{orange}{(+0.6)} & \textcolor{blue}{(-22.6\%)} & \textcolor{orange}{(+1.6)} & \textcolor{blue}{(-22.3\%)} & \textcolor{orange}{(+6.9)} & \textcolor{blue}{(-12.0\%)} & \textcolor{orange}{(+0.8)} & \textcolor{blue}{(-11.8\%)} & \textcolor{orange}{(+4.3)} & \textcolor{blue}{(-17.1\%)} \\
\midrule
\multicolumn{13}{l}{\textbf{Qwen3-14B}} \\
Vanilla & 95.6 & 1433 & 96.2 & 4559 & 96.9 & 7277 & 67.1 & 16019 & \underline{90.5} & 7252 & \underline{63.6} & 7323 \\
CCoT & 95.8 & \textbf{969} & 96.5 & 3623 & \underline{97.2} & \underline{5852} & 69.6 & 14349 & 90.1 & 6465 & 62.1 & 6530 \\
DEER & \underline{96.0} & 1051 & 94.4 & \textbf{3262} & 94.8 & 6065 & 65.8 & 13983 & 85.3 & \textbf{5687} & 61.7 & 6971 \\
SEAL & 93.6 & 1181 & 95.4 & 3788 & 96.3 & 6193 & 61.7 & 14714 & 88.7 & 6411 & 62.9 & \underline{6421} \\
STIR & 95.9 & 1197 & 92.4 & 3825 & 95.3 & 6295 & 67.7 & \underline{13239} & 86.8 & 6680 & 62.6 & 6923 \\
ReBalance & 95.6 & 1740 & 95.4 & 4222 & 97.0 & 5999 & \textbf{71.0} & 15340 & 89.9 & 6312 & 63.1 & 6528 \\
Ours & \textbf{96.4} & \underline{1011} & \textbf{96.8} & \underline{3510} & \textbf{98.3} & \textbf{5751} & \underline{70.4} & \textbf{13144} & \textbf{91.0} & \underline{6210} & \textbf{64.1} & \textbf{6309} \\
$\Delta$ vs. Van. & \textcolor{orange}{(+0.8)} & \textcolor{blue}{(-29.4\%)} & \textcolor{orange}{(+0.6)} & \textcolor{blue}{(-23.0\%)} & \textcolor{orange}{(+1.4)} & \textcolor{blue}{(-21.0\%)} & \textcolor{orange}{(+3.3)} & \textcolor{blue}{(-17.9\%)} & \textcolor{orange}{(+0.5)} & \textcolor{blue}{(-14.4\%)} & \textcolor{orange}{(+0.5)} & \textcolor{blue}{(-13.8\%)} \\
\bottomrule
\end{tabular}
}
\end{table}

\subsection{Experimental Setups}

\textbf{Models and Datasets.} We evaluate the effectiveness of our proposed method across six diverse benchmarks. These include four mathematical reasoning datasets: GSM8K \citep{cobbe2021training}, MATH500 \citep{hendrycks2measuring}, AMC 2023 \citep{amc2023}, and AIME 2025 \citep{aime2025}; one scientific reasoning dataset: GPQA Diamond \citep{rein2024gpqa}; and one code reasoning dataset: LiveCodeBench \citep{jainlivecodebench}. Model performance is measured using Accuracy (\textbf{Acc}, $\uparrow$) and the average number of generated tokens (\textbf{Tok}, $\downarrow$). Given the limited number of samples in datasets AMC 2023 and AIME 2025, we conduct 16 sampling rounds per instance to ensure evaluation stability. We report the mean performance in the main text and the standard deviations in Appendix \ref{app:statistical_reliability}. We evaluate our approach using widely adopted LRMs, including DeepSeek-R1-Distill-Qwen-1.5B/7B \citep{guo2025deepseek} and Qwen3-4B/14B \citep{qwen3technicalreport}.

\textbf{Baselines.} We compare our approach against six training-free baselines. \textbf{Vanilla} represents the direct evaluation of the model without any intervention. The remaining methods can be categorized as follows. For prompt-based methods, we compare against \textbf{CCoT} \citep{renze2024benefits}, which explicitly instructs the model to think concisely. For early-exit methods, we compare against \textbf{DEER} \citep{yang2025dynamic}, which self-truncate CoT sequences by early exit during generation. Most importantly, because our proposed approach is closely related to steering-based methods, we select three representative methods from this category for a comprehensive comparison: \textbf{SEAL} \citep{chenseal}, \textbf{STIR} \citep{shi2026internalizing}, and \textbf{ReBalance} \citep{liefficient}. Specifically, SEAL computes a steering vector to calibrate the CoT process, guiding the model toward more concise reasoning. STIR formulates reasoning enhancement as a dynamic latent trajectory control problem, injecting context-specific steering vector via a library of self-distilled internal reasoning tools. ReBalance utilizes real-time confidence to modulate a steering vector, pruning redundancy during overthinking and promoting exploration during underthinking.

\textbf{Implementation Details.} Our implementations are built upon a modified version of the Easysteer \citep{xu2025easysteer} library to integrate the projection operation, utilizing the vLLM \citep{kwon2023efficient} backend for high-throughput inference. For the QP solving, we precompute an mp-QP via the PDAQP solver \citep{arnstrom2024pdaqp}. During the inference phase, solving the QP is transformed into a region querying and affine mapping process, implemented using PyTorch operators. Detailed hyperparameter and decoding settings are provided in Appendix \ref{app:experment_setting}. For baseline methods, we use their official implementations and the default settings. All experiments were conducted on a server equipped with eight 80GB NVIDIA A800 GPUs.

\subsection{Main Results}
\textbf{Performance on Mathematical Reasoning.} As shown in Table 1, our proposed method consistently outperforms the other baselines across all four mathematical benchmarks and various model scales, successfully reducing token counts while improving accuracy. We observe that baseline methods such as DEER and CCoT yield lower token counts on some datasets. For DEER, while it uses fewer tokens on MATH500 for the 1.5B model, its aggressive truncation causes a catastrophic 9.9\% drop in accuracy, indicating severe underthinking that skips necessary logical deductions. Another baseline, CCoT, achieves lower token generation on simpler tasks like GSM8K. One reason is that GSM8K requires relatively shallow reasoning, allowing explicit instructions to compress the output. However, CCoT's performance degrades noticeably on more complex benchmarks. 

\textbf{Cross-Domain Generalization.} To investigate the generalizability of our method beyond mathematical domains, we extended our evaluation to code generation (LiveCodeBench) and scientific knowledge reasoning (GPQA-Diamond). As detailed in Table 1, our method maintains its strong performance in these distinct domains. These results demonstrate that our method can generalize well to improve generation efficiency and accuracy across diverse reasoning tasks.

\textbf{Further Supplementary Evaluations.} Due to space limitations, we provide extensive supplementary experiments in Appendix \ref{app:add_exp}. First, we demonstrate the cross-domain and cross-difficulty transferability of our method in Table \ref{tab:combined_performance}, showing that efficient reasoning sets derived from basic math or science domains can successfully generalize to complex competition-level math and coding tasks. Second, our semantic change analysis in Table \ref{tab:semantic_analysis} reveals that the proposed method effectively suppresses redundant self-verification (Reflection) and perspective-shifting (Transition) behaviors. To further investigate whether the proposed method introduces any unintended side effects on general capabilities, we evaluate our method on the Creative Writing v3 benchmark in Table \ref{tab:creative_writing}, demonstrating that the model's creative expression are preserved. Moreover, we conduct a Pass@k performance analysis on AMC23 and AIME2025 to investigate whether our method affects the model's exploration capability. The results in Table \ref{tab:pass_k_performance} indicates that our approach maintains or improves the Pass@k performance.

\begin{figure}[h]
    \centering
    \includegraphics[width=\textwidth]{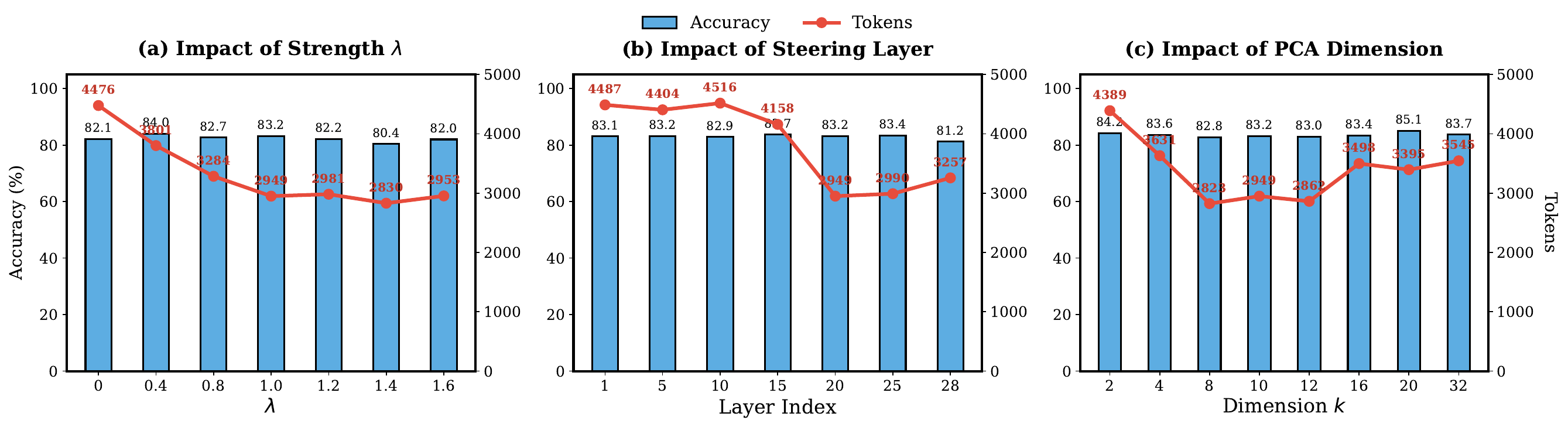} 
    \caption{The impact of the key hyperparameters for performance in MATH500: (a) Intervention Strength. (b) Steering Layer. (c) PCA dimension.}
    \label{fig:ablation_studies}
\end{figure}

\subsection{Quantitative Evaluation of Efficiency} \label{sec:efficiency}
To validate the efficiency of our GPU-accelerated projection (\textbf{Explicit}), we compare it against an \textbf{Online} baseline using the widely adopted CvxpyLayer solver \citep{cvxpylayers2019} to solve QP. We evaluate both settings across two dimensions: the computational overhead and the inference latency in vLLM.

\begin{wraptable}{r}{0.5\columnwidth} 
\vspace{1em} 
\centering
\caption{Efficiency Analysis on A800 GPU.}
\label{tab:performance_comparison}
\resizebox{\linewidth}{!}{
\begin{tabular}{lccc}
\toprule
\textbf{Method} & \textbf{\# Tokens} & \textbf{TPS $\uparrow$} & \textbf{TPR (s) $\downarrow$} \\
\midrule
\multicolumn{4}{l}{\textit{DeepSeek-R1-Distill-Qwen-1.5B}} \\
Baseline        & 4678 & 5596 & 0.83 \\
Ours (Online)   & 3074 & 4682 & 0.66 (-21.2\%) \\
Ours (Explicit) & 2812 & 5472 & 0.51 (-38.4\%) \\
\midrule
\multicolumn{4}{l}{\textit{DeepSeek-R1-Distill-Qwen-7B}} \\
Baseline        & 3631 & 2659 & 1.36 \\
Ours (Online)   & 2569 & 2262 & 1.14 (-16.2\%) \\
Ours (Explicit) & 2685 & 2638 & 1.02 (-25.0\%) \\
\midrule
\multicolumn{4}{l}{\textit{Qwen3-4B}} \\
Baseline        & 4593 & 1242 & 3.70 \\
Ours (Online)   & 3608 & 1178 & 3.06 (-17.3\%) \\
Ours (Explicit) & 3504 & 1339 & 2.62 (-29.2\%) \\
\midrule
\multicolumn{4}{l}{\textit{Qwen3-14B}} \\
Baseline        & 4337 & 786 & 5.51 \\
Ours (Online)   & 3520 & 719 & 4.88 (-11.4\%) \\
Ours (Explicit) & 3484 & 842 & 4.14 (-24.9\%) \\
\bottomrule
\end{tabular}
}
\vspace{-0.5em} 
\end{wraptable}

\textbf{Computational Overhead Analysis.} Table \ref{tab:solving_time} presents the detailed computational cost comparison for solving a single QP instance. As the results demonstrate, the computational overhead introduced by our \textbf{Explicit} method is exceptionally small. While the \textbf{Online} solver requires a few milliseconds, our GPU acceleration drastically reduces this solving time to less than a millisecond (e.g., 0.3 ms for the 1.5B model), requiring less than 2 MB of additional GPU memory across all evaluated models. A detailed analysis on the computational overhead is provided in Appendix \ref{app:overhead}.

\textbf{Efficiency Analysis.} Table \ref{tab:performance_comparison} presents the end-to-end efficiency results against native \texttt{vLLM} on MATH500 using NVIDIA A800 GPUs. We generate 3 responses per problem and evaluate the end-to-end inference latency using Time Per Request (TPR). Our method consistently achieves lower TPR across all models compared to the baseline, an acceleration primarily driven by the drastic reduction in generated tokens. Notably, our \textbf{Explicit} version achieves a Tokens Per Second (TPS) rate highly comparable to the native baseline. These results strongly demonstrate the practicability of our approach for high-throughput, real-world deployment. The slight token variations between \textbf{Ours (Online)} and \textbf{Ours (Explicit)} stem from stochastic decoding.

\subsection{Ablation Study} \label{sec:ablation}
In this subsection, we conduct ablation studies on the MATH500 dataset using the DeepSeek-R1-Distill-Qwen-1.5B model to evaluate the impact of the key hyperparameters in our methods. See Appendix \ref{app:ablation} for more details about ablation study of other models.

\textbf{Impact of Intervention Strength $\lambda$.} 
Figure \ref{fig:ablation_studies}(a) shows that $\lambda$ in $[1.0, 1.6]$ yield substantial token reduction. However, we observe a slight trade-off where increasing $\lambda$ leads to a marginal decrease in accuracy. This means that the default choice of $\lambda = 1.0$ achieves an appropriate balance.

\textbf{Impact of Steering Layer.} 
Figure \ref{fig:ablation_studies}(b) illustrates the sensitivity of steering performance to the chosen intervention layer. Our results indicate that middle-to-deep layers yield the better performance. This observation can be explained by the typical forward pass dynamics of LLMs. Intervening at shallow layers is ineffective because these layers primarily process token-level representations rather than abstract reasoning features. Conversely, corrections at the final layers yield diminishing returns, as these representations are to summarize the context for next-token prediction. 

\textbf{Impact of PCA Dimension $k$.} 
Figure \ref{fig:ablation_studies}(c) demonstrates the effect of the latent space dimension $k$. Low dimensions lead to information loss, making it difficult to construct a useful set $\mathcal{H}$. In contrast, very high dimensions introduce unnecessary noise that hinders the effectiveness of the projection. The performance remains stable and efficient when $k$ is within the range of $[8, 16]$.

\textbf{Impact of Cluster Number $M$.}
Figure \ref{fig:ablation_studies_cluster} shows that utilizing a small number of clusters (e.g., $M \le 4$) yields marginal token reduction. This limited efficiency gain occurs because fewer linear constraints form an inaccurate and coarse convex polytope as $\mathcal{H}$. As $M$ increases, the efficient reasoning set boundary becomes more precise, resulting in stabilized token reduction and stronger efficiency gains. However, a larger $M$ introduces more constraints to the QP problem, which correspondingly increases the solving time during inference.
 
\textbf{Impact of Boundary Proportion $\alpha$.} 
As illustrated in Figure \ref{fig:ablation_studies_coverage}, adjusting $\alpha$ within the range of 20\% to 100\% consistently achieves notable token reduction. However, smaller values of $\alpha$ lead to a slight degradation in model accuracy. This performance drop occurs because a low $\alpha$ creates a highly restrictive set, causing aggressive interventions that may inadvertently discard useful semantic variations. Conversely, when $\alpha = 100\%$, the token reduction becomes less pronounced. This is likely because a fully inclusive set $\mathcal{H}$ may also encompass "overthinking" behaviors.

\section{Conclusion}
In this work, we investigated the phenomenon of overthinking and underthinking in LRMs from a geometric perspective. By analyzing the reasoning steps in latent space, we identified that efficient reasoning steps naturally form a concentrated region, while those deviating from this region tend to produce verbose sequence. Based on this observation, we proposed a training-free method that utilizes quadratic programming to steer the model's hidden states back into the region during the inference. Our empirical results demonstrate that the proposed method can improve reasoning performance while reducing token generation costs. 

\section{Acknowledgements}
This work was supported by National Natural Science Foundation of China (62325305), and the BNRist project
(No. BNR2024TD03003). We thank Yee Hin Chong from Tsinghua University for supporting this work with GPU resources.

\bibliographystyle{unsrt} 
\bibliography{main}

@article{novikov2025alphaevolve,
  title={Alphaevolve: A coding agent for scientific and algorithmic discovery},
  author={Novikov, Alexander and V{\~u}, Ng{\^a}n and Eisenberger, Marvin and Dupont, Emilien and Huang, Po-Sen and Wagner, Adam Zsolt and Shirobokov, Sergey and Kozlovskii, Borislav and Ruiz, Francisco JR and Mehrabian, Abbas and others},
  journal={arXiv preprint arXiv:2506.13131},
  year={2025}
}

@article{wei2022chain,
  title={Chain-of-thought prompting elicits reasoning in large language models},
  author={Wei, Jason and Wang, Xuezhi and Schuurmans, Dale and Bosma, Maarten and Xia, Fei and Chi, Ed and Le, Quoc V and Zhou, Denny and others},
  journal={Advances in neural information processing systems},
  volume={35},
  pages={24824--24837},
  year={2022}
}

@article{guo2025deepseek,
  title={Deepseek-r1: Incentivizing reasoning capability in llms via reinforcement learning},
  author={Guo, Daya and Yang, Dejian and Zhang, Haowei and Song, Junxiao and Wang, Peiyi and Zhu, Qihao and Xu, Runxin and Zhang, Ruoyu and Ma, Shirong and Bi, Xiao and others},
  journal={arXiv preprint arXiv:2501.12948},
  year={2025}
}

@article{wu2025more,
  title={When more is less: Understanding chain-of-thought length in llms},
  author={Wu, Yuyang and Wang, Yifei and Ye, Ziyu and Du, Tianqi and Jegelka, Stefanie and Wang, Yisen},
  journal={arXiv preprint arXiv:2502.07266},
  year={2025}
}

@article{huang2025adactrl,
  title={AdaCtrl: Towards Adaptive and Controllable Reasoning via Difficulty-Aware Budgeting},
  author={Huang, Shijue and Wang, Hongru and Zhong, Wanjun and Su, Zhaochen and Feng, Jiazhan and Cao, Bowen and Fung, Yi R},
  journal={arXiv preprint arXiv:2505.18822},
  year={2025}
}

@inproceedings{li2026making,
  title={Making slow thinking faster: Compressing LLM chain-of-thought via step entropy},
  author={Li, Zeju and Zhong, Jianyuan and Zheng, Ziyang and Wen, Xiangyu and Xu, Zhijian and Cheng, Yingying and Zhang, Fan and Xu, Qiang},
  booktitle={The Fourteenth International Conference on Learning Representations},
  year={2026}
}

@article{yizhar2011neocortical,
  title={Neocortical excitation/inhibition balance in information processing and social dysfunction},
  author={Yizhar, Ofer and Fenno, Lief E and Prigge, Matthias and Schneider, Franziska and Davidson, Thomas J and O’shea, Daniel J and Sohal, Vikaas S and Goshen, Inbal and Finkelstein, Joel and Paz, Jeanne T and others},
  journal={Nature},
  volume={477},
  number={7363},
  pages={171--178},
  year={2011},
  publisher={Nature Publishing Group UK London}
}

@article{cobbe2021training,
  title={Training verifiers to solve math word problems},
  author={Cobbe, Karl and Kosaraju, Vineet and Bavarian, Mohammad and Chen, Mark and Jun, Heewoo and Kaiser, Lukasz and Plappert, Matthias and Tworek, Jerry and Hilton, Jacob and Nakano, Reiichiro and others},
  journal={arXiv preprint arXiv:2110.14168},
  year={2021}
}

@inproceedings{hendrycks2measuring,
  title={Measuring Mathematical Problem Solving With the MATH Dataset},
  author={Hendrycks, Dan and Burns, Collin and Kadavath, Saurav and Arora, Akul and Basart, Steven and Tang, Eric and Song, Dawn and Steinhardt, Jacob},
  year = {2021},
  booktitle={Thirty-fifth Conference on Neural Information Processing Systems Datasets and Benchmarks Track (Round 2)}
}

@misc{amc2023,
  author = {{Mathematical Association of America}},
  title = {American Mathematics Competitions ({AMC}) 10/12},
  year = {2023},
  howpublished = {Problems and Answer Keys}
}

@misc{aime2025,
  author = {{Mathematical Association of America}},
  title = {American Invitational Mathematics Examination ({AIME})},
  year = {2025},
  howpublished = {Problems and Answer Keys}
}

@inproceedings{rein2024gpqa,
  title={Gpqa: A graduate-level google-proof q\&a benchmark},
  author={Rein, David and Hou, Betty Li and Stickland, Asa Cooper and Petty, Jackson and Pang, Richard Yuanzhe and Dirani, Julien and Michael, Julian and Bowman, Samuel R},
  booktitle={First conference on language modeling},
  year={2024}
}

@inproceedings{jainlivecodebench,
  title={LiveCodeBench: Holistic and Contamination Free Evaluation of Large Language Models for Code},
  author={Jain, Naman and Han, King and Gu, Alex and Li, Wen-Ding and Yan, Fanjia and Zhang, Tianjun and Wang, Sida and Solar-Lezama, Armando and Sen, Koushik and Stoica, Ion},
  year = {2025},
  booktitle={The Thirteenth International Conference on Learning Representations}
}

@misc{qwen3technicalreport,
      title={Qwen3 Technical Report}, 
      author={Qwen Team},
      year={2025},
      eprint={2505.09388},
      archivePrefix={arXiv},
      primaryClass={cs.CL},
      url={https://arxiv.org/abs/2505.09388}, 
}

@inproceedings{renze2024benefits,
  title={The Benefits of a Concise Chain of Thought on Problem-Solving in Large Language Models},
  author={Renze, Matthew and Guven, Erhan},
  booktitle={2024 2nd International Conference on Foundation and Large Language Models (FLLM)},
  pages={476--483},
  year={2024},
  organization={IEEE}
}

@article{yang2025dynamic,
  title={Dynamic early exit in reasoning models},
  author={Yang, Chenxu and Si, Qingyi and Duan, Yongjie and Zhu, Zheliang and Zhu, Chenyu and Li, Qiaowei and Chen, Minghui and Lin, Zheng and Wang, Weiping},
  journal={arXiv preprint arXiv:2504.15895},
  year={2025}
}

@inproceedings{chenseal,
  title={SEAL: Steerable Reasoning Calibration of Large Language Models for Free},
  author={Chen, Runjin and Zhang, Zhenyu and Hong, Junyuan and Kundu, Souvik and Wang, Zhangyang},
  booktitle={Second Conference on Language Modeling},
  year={2025}
}

@article{shi2026internalizing,
  title={Internalizing LLM Reasoning via Discovery and Replay of Latent Actions},
  author={Shi, Zhenning and Zhu, Yijia and Shi, Junhan and Zhang, Xun and Wang, Lei and Miao, Congcong},
  journal={arXiv preprint arXiv:2602.04925},
  year={2026}
}

@inproceedings{liefficient,
  title={Efficient Reasoning with Balanced Thinking},
  author={Li, Yulin and Tu, Tengyao and Ding, Li and Wang, Junjie and Zhen, Hui-Ling and Chen, Yixin and Li, Yong and Tian, Zhuotao},
  booktitle={The Fourteenth International Conference on Learning Representations},
  year={2026}
}

@article{xu2025easysteer,
  title={Easysteer: A unified framework for high-performance and extensible llm steering},
  author={Xu, Haolei and Mei, Xinyu and Yan, Yuchen and Zhou, Rui and Zhang, Wenqi and Lu, Weiming and Zhuang, Yueting and Shen, Yongliang},
  journal={arXiv preprint arXiv:2509.25175},
  year={2025}
}

@inproceedings{kwon2023efficient,
  title={Efficient memory management for large language model serving with pagedattention},
  author={Kwon, Woosuk and Li, Zhuohan and Zhuang, Siyuan and Sheng, Ying and Zheng, Lianmin and Yu, Cody Hao and Gonzalez, Joseph and Zhang, Hao and Stoica, Ion},
  booktitle={Proceedings of the 29th symposium on operating systems principles},
  pages={611--626},
  year={2023}
}

@inproceedings{cvxpylayers2019,
  author={Agrawal, A. and Amos, B. and Barratt, S. and Boyd, S. and Diamond, S. and Kolter, Z.},
  title={Differentiable Convex Optimization Layers},
  booktitle={Advances in Neural Information Processing Systems},
  year={2019},
}

@inproceedings{arnstrom2024pdaqp,
  author={Arnström, Daniel and Axehill, Daniel},
  booktitle={2024 IEEE 63rd Conference on Decision and Control (CDC)}, 
  title={A High-Performant Multi-Parametric Quadratic Programming Solver}, 
  year={2024},
  volume={},
  number={},
  pages={303-308},
}

@article{karan2025reasoning,
  title={Reasoning with sampling: Your base model is smarter than you think},
  author={Karan, Aayush and Du, Yilun},
  journal={arXiv preprint arXiv:2510.14901},
  year={2025}
}

@article{bemporad2002explicit,
  title={The explicit linear quadratic regulator for constrained systems},
  author={Bemporad, Alberto and Morari, Manfred and Dua, Vivek and Pistikopoulos, Efstratios N},
  journal={Automatica},
  volume={38},
  number={1},
  pages={3--20},
  year={2002},
  publisher={Elsevier}
}

@article{wang2025wait,
  title={Wait, we don’t need to" wait"! removing thinking tokens improves reasoning efficiency},
  author={Wang, Chenlong and Feng, Yuanning and Chen, Dongping and Chu, Zhaoyang and Krishna, Ranjay and Zhou, Tianyi}
}

@inproceedings{lincontrolling,
  title={Controlling Thinking Speed in Reasoning Models},
  author={Lin, Zhengkai and Fu, Zhihang and Chen, Ze and Chen, Chao and Xie, Liang and Wang, Wenxiao and Cai, Deng and Wang, Zheng and Ye, Jieping},
  booktitle={The Thirty-ninth Annual Conference on Neural Information Processing Systems}
}

@inproceedings{haotraining,
  title={Training Large Language Models to Reason in a Continuous Latent Space},
  author={Hao, Shibo and Sukhbaatar, Sainbayar and Su, DiJia and Li, Xian and Hu, Zhiting and Weston, Jason E and Tian, Yuandong},
  booktitle={Second Conference on Language Modeling}
}

@inproceedings{skean2025layer,
  title={Layer by Layer: Uncovering Hidden Representations in Language Models},
  author={Skean, Oscar and Arefin, Md Rifat and Zhao, Dan and Patel, Niket Nikul and Naghiyev, Jalal and Lecun, Yann and Shwartz-Ziv, Ravid},
  booktitle={International Conference on Machine Learning},
  pages={55854--55875},
  year={2025},
  organization={PMLR}
}

@inproceedings{tack2025llm,
  title={LLM Pretraining with Continuous Concepts},
  author={Tack, Jihoon and Lanchantin, Jack and Yu, Jane and Cohen, Andrew and Kulikov, Ilia and Lan, Janice and Hao, Shibo and Tian, Yuandong and Weston, Jason E and Li, Xian},
  booktitle={Mechanistic Interpretability Workshop at NeurIPS 2025}
}

@article{gozeten2025continuous,
  title={Continuous chain of thought enables parallel exploration and reasoning},
  author={Gozeten, Halil Alperen and Ildiz, M Emrullah and Zhang, Xuechen and Harutyunyan, Hrayr and Rawat, Ankit Singh and Oymak, Samet},
  journal={arXiv preprint arXiv:2505.23648},
  year={2025}
}

@inproceedings{yang2024large,
  title={Do large language models latently perform multi-hop reasoning?},
  author={Yang, Sohee and Gribovskaya, Elena and Kassner, Nora and Geva, Mor and Riedel, Sebastian},
  booktitle={Proceedings of the 62nd Annual Meeting of the Association for Computational Linguistics (Volume 1: Long Papers)},
  pages={10210--10229},
  year={2024}
}

@article{zhang2025soft,
  title={Soft thinking: Unlocking the reasoning potential of llms in continuous concept space},
  author={Zhang, Zhen and He, Xuehai and Yan, Weixiang and Shen, Ao and Zhao, Chenyang and Wang, Shuohang and Shen, Yelong and Wang, Xin Eric},
  journal={arXiv preprint arXiv:2505.15778},
  year={2025}
}

@article{li2025cot,
  title={CoT Vectors: Transferring and Probing the Reasoning Mechanisms of LLMs},
  author={Li, Li and Wang, Ziyi and Wu, Yongliang and Cai, Jianfei and Yang, Xu},
  journal={arXiv preprint arXiv:2510.00579},
  year={2025}
}

@article{cheng2024compressed,
  title={Compressed chain of thought: Efficient reasoning through dense representations},
  author={Cheng, Jeffrey and Van Durme, Benjamin},
  journal={arXiv preprint arXiv:2412.13171},
  year={2024}
}

@article{huang2025mitigating,
  title={Mitigating overthinking in large reasoning models via manifold steering},
  author={Huang, Yao and Chen, Huanran and Ruan, Shouwei and Zhang, Yichi and Wei, Xingxing and Dong, Yinpeng},
  journal={arXiv preprint arXiv:2505.22411},
  year={2025}
}

@article{nguyen2026atlas,
  title={ATLAS: Adaptive Test-Time Latent Steering with External Verifiers for Enhancing LLMs Reasoning},
  author={Nguyen, Tuc and Le, Thai},
  journal={arXiv preprint arXiv:2601.03093},
  year={2026}
}

@misc{creative-writing-bench-v3,
  author = {Samuel J Paech},
  title = {EQ-Bench Creative Writing Benchmark v3},
  year = {2025},
  publisher = {GitHub},
  journal = {GitHub repository},
  howpublished = {\url{https://github.com/EQ-bench/creative-writing-bench}}
}

@inproceedings{qiao2025concise,
  title={Concise: Confidence-guided compression in step-by-step efficient reasoning},
  author={Qiao, Ziqing and Deng, Yongheng and Zeng, Jiali and Wang, Dong and Wei, Lai and Wang, Guanbo and Meng, Fandong and Zhou, Jie and Ren, Ju and Zhang, Yaoxue},
  booktitle={Proceedings of the 2025 Conference on Empirical Methods in Natural Language Processing},
  pages={8021--8040},
  year={2025}
}

@article{zhang2025reasoning,
  title={Reasoning Models Know When They're Right: Probing Hidden States for Self-Verification},
  author={Zhang, Anqi and Chen, Yulin and Pan, Jane and Zhao, Chen and Panda, Aurojit and Li, Jinyang and He, He},
  journal={arXiv preprint arXiv:2504.05419},
  year={2025}
}

@article{yu2025long,
  title={Long-short chain-of-thought mixture supervised fine-tuning eliciting efficient reasoning in large language models},
  author={Yu, Bin and Yuan, Hang and Li, Haotian and Xu, Xueyin and Wei, Yuliang and Wang, Bailing and Qi, Weizhen and Chen, Kai},
  journal={arXiv preprint arXiv:2505.03469},
  year={2025}
}

@article{yuan2025not,
  title={Not all tokens are what you need in thinking},
  author={Yuan, Hang and Yu, Bin and Li, Haotian and Yang, Shijun and Wang, Christina Dan and Yu, Zhou and Xu, Xueyin and Qi, Weizhen and Chen, Kai},
  journal={arXiv preprint arXiv:2505.17827},
  year={2025}
}

@article{jiang2025think,
  title={Think only when you need with large hybrid-reasoning models},
  author={Jiang, Lingjie and Wu, Xun and Huang, Shaohan and Dong, Qingxiu and Chi, Zewen and Dong, Li and Zhang, Xingxing and Lv, Tengchao and Cui, Lei and Wei, Furu},
  journal={arXiv preprint arXiv:2505.14631},
  year={2025}
}

@article{luo2025autol2s,
  title={Autol2s: Auto long-short reasoning for efficient large language models},
  author={Luo, Feng and Chuang, Yu-Neng and Wang, Guanchu and Le, Hoang Anh Duy and Zhong, Shaochen and Liu, Hongyi and Yuan, Jiayi and Sui, Yang and Braverman, Vladimir and Chaudhary, Vipin and others},
  journal={arXiv preprint arXiv:2505.22662},
  year={2025}
}

@inproceedings{liu2024fantastic,
  title={Fantastic semantics and where to find them: Investigating which layers of generative llms reflect lexical semantics},
  author={Liu, Zhu and Kong, Cunliang and Liu, Ying and Sun, Maosong},
  booktitle={Findings of the Association for Computational Linguistics: ACL 2024},
  pages={14551--14558},
  year={2024}
}

@article{arnstrom2022dual,
	title={A dual active-set solver for embedded quadratic programming using recursive {$LDL^T$} updates},
	author={Arnstr{\"o}m, Daniel and Bemporad, Alberto and Axehill, Daniel},
	journal={IEEE Transactions on Automatic Control},
	volume={67},
	number={8},
	pages={4362--4369},
	year={2022},
	publisher={IEEE}
}

@article{stellato2020osqp,
	title={{OSQP}: An operator splitting solver for quadratic programs},
	author={Stellato, Bartolomeo and Banjac, Goran and Goulart, Paul and Bemporad, Alberto and Boyd, Stephen},
	journal={Mathematical Programming Computation},
	volume={12},
	number={4},
	pages={637--672},
	year={2020},
	publisher={Springer}
}


\appendix
\newpage
\section*{Appendix Table of Contents}
\startcontents[appendix]
\printcontents[appendix]{l}{1}{\setcounter{tocdepth}{2}}
\newpage

\section{Related Work Details} \label{app:related_work}
\textbf{Implicit Reasoning in Latent Space.} Recent studies reveal that large language models inherently perform latent reasoning within their hidden computations \citep{haotraining, skean2025layer, tack2025llm}. For instance, empirical investigations show that multiple parallel latent reasoning paths can coexist within the middle layers of transformer architectures \citep{gozeten2025continuous}, and that intermediate entities in multi-hop reasoning tasks can be directly recovered from these latent representations \citep{yang2024large}. Collectively, these observations have catalyzed a surge of research into implicit efficient reasoning within the latent space. To harness this capability, ``soft-thinking'' approaches replace discrete thinking tokens with continuous concept tokens \citep{zhang2025soft}. Another paradigm directly trains model to reason in a continuous latent space \citep{haotraining}. Additionally, several studies focus on compressing long, explicit CoT trajectories into dense vectors to reduce inference overhead \citep{li2025cot, cheng2024compressed}. To further explore efficient reasoning in the latent space, our study reveals that efficient reasoning steps form a concentrated region in the latent space. Motivated by this, we formulate a quadratic program to keep the hidden state of the model focus within this region, implicitly guiding the model to generate more effective reasoning steps.

\textbf{Mitigating Overthinking in LLMs.} To alleviate the excessive computational burden of overthinking, recent post-training approaches aim to reshape the model's reliance on lengthy reasoning. These include supervised fine-tuning methods that utilize compressed CoT data \citep{yuan2025not, yu2025long}, as well as reinforcement learning-based methods that optimize the accuracy-efficiency trade-off through reward shaping \citep{luo2025autol2s, jiang2025think}. While empirically effective, these methods incur significant retraining costs.

Consequently, training-free mechanisms have gained substantial traction. One paradigm is \textit{Early Exiting}, which halts the reasoning process once sufficient evidence is gathered, relying on either external monitors \citep{zhang2025reasoning} or internal confidence signals \citep{yang2025dynamic, qiao2025concise}. However, these termination decisions are often coarse-grained and risk discarding potentially valuable reasoning steps, inevitably inducing underthinking. Another promising training-free paradigm is \textit{Representation Steering}. This approach intervenes directly within the model's latent activation space during inference to steer the reasoning trajectory and suppress redundant loops \citep{huang2025mitigating, chenseal}. Our work naturally falls into this category. 

\textbf{Efficient Reasoning via Representation Steering.} Representation steering has attracted growing interest for enhancing reasoning efficiency, motivated by its training-free nature and mechanistic interpretability \citep{huang2025mitigating, chenseal, nguyen2026atlas}. These methods first identify linear directions within the activation space that correspond to specific model behaviors. By steering these linear directions during inference, they can calibrate model output, such as suppressing redundant loops and meaningless verbosity to mitigate overthinking. However, steering techniques typically apply a static, unidirectional intervention, which may not capture the highly dynamic nature of a reasoning trajectory \citep{shi2026internalizing}. To address this, recent studies have developed adaptive mechanisms. For example, ReBalance adjusts steering direction and strength based on real-time confidence assessments \citep{liefficient}, while STIR operates as a dynamic controller that dictates intervention timing through latent trajectory monitoring \citep{shi2026internalizing}. 

While empirically effective, these approaches above are reactive: they focus on pushing the latent state away from problematic behaviors, lacking a principled consideration of \textit{where the efficient reasoning state should actually reside.} This distinction is particularly critical in the context of efficient reasoning; if the intervention merely repels the state from overthinking without a defined target, it inevitably risks over-correction and induces underthinking. To address this gap, our work reveals the geometric concentration of efficient reasoning within the latent space. We explicitly locate this concentrated region and formulate representation steering as a constrained optimization problem, directly projecting deviant hidden states into this region for correction.

\section{More Discussions about Progress}
\subsection{Property of Progress} \label{app:prop_progress}

In this subsection, we investigate the mathematical relationship between the step-wise progress and the performance (accuracy and sequence length) of the entire reasoning trajectory.

\begin{property}[Cumulative Progress]
\label{prop:cumulative_progress}
\textit{Given a question $q$ and a generated reasoning trajectory $c = [c_{1}, c_{2}, \dots, c_{N}]$, the cumulative sum of step-wise progress determines the performance of final generation relative to the baseline. Specifically:}
\begin{equation} \label{prop:acc}
\mathbb{P}_{y \sim \pi(\cdot | c, q)}(y=y^*) = \mathbb{P}_{(c, y) \sim \pi( \cdot | q)}(y=y^*) + \sum_{t=1}^{N} p_\text{acc}(c_{t}; c_{<t}, q)
\end{equation}

\begin{equation} \label{prop:length}
\sum_{t=1}^N Len(c_t) + \mathbb{E}_{y \sim \pi(\cdot | c, q)}[Len(y)] = \mathbb{E}_{(c, y) \sim \pi(\cdot | q)}[Len(c, y)] - \sum_{t=1}^{N}p_\text{len}(c_{t}; c_{<t}, q)
\end{equation}
\end{property}

\begin{proof}

\textbf{Proof for Eq. \eqref{prop:acc}.} Let $P_{t} = \mathbb{P}_{(c_{>t},y)\sim\pi(\cdot \mid c_{\le t},q)}(y=y^{*})$ denote the expected accuracy after observing the partial trajectory up to step $t$. By definition, the step-wise accuracy progress is $p_\text{acc}(c_{t}) = P_{t} - P_{t-1}$. Summing this progress over all $N$ steps yields a telescoping series:
\begin{equation} \notag
        \sum_{t=1}^{N} p_\text{acc}(c_{t}) = \sum_{t=1}^{N} (P_{t} - P_{t-1}) = P_{N} - P_{0}
\end{equation}
Note the boundary conditions: $P_{0} = \mathbb{P}_{(c, y) \sim \pi( \cdot \mid q)}(y=y^{*})$ corresponds to the baseline expected accuracy before any reasoning steps are generated, and $P_{N} = \mathbb{P}_{y \sim \pi(\cdot \mid c, q)}(y=y^*)$ represents the realized correctness probability given the full generated trajectory $c = c_{\le N}$. Rearranging the equation $P_{N} - P_{0} = \sum_{t=1}^{N} p_\text{acc}(c_{t})$ directly yields Eq. \eqref{prop:acc}.

\textbf{Proof for Eq. \eqref{prop:length}.} Similarly, let $L_{t} = \mathbb{E}_{(c_{> t},y)\sim\pi(\cdot \mid c_{\le t},q)}[Len(c_{> t}, y)]$ denote the expected remaining sequence length after generating step $t$. The step-wise progress for length is defined as $p_\text{len}(c_{t}) = L_{t-1} - L_{t} - Len(c_t)$. Summing $p_\text{len}(c_t)$ from $t=1$ to $N$ gives:
\begin{equation} \notag
    \sum_{t=1}^{N} p_\text{len}(c_{t}) = \sum_{t=1}^{N} \Big( L_{t-1} - L_{t} - Len(c_{t}) \Big) = (L_{0} - L_{N}) - \sum_{t=1}^{N} Len(c_{t})
\end{equation}
By examining the boundary conditions, we have $L_{0} = \mathbb{E}_{(c, y) \sim \pi(\cdot \mid q)}[Len(c, y)]$, which is the total expected length of a reasoning trajectory from the initial question $q$. Meanwhile, $L_{N} = \mathbb{E}_{y \sim \pi(\cdot \mid c, q)}[Len(y)]$ is the expected length of the final answer $y$ after the complete reasoning trajectory $c$ has been generated. Substituting $L_{0}$ and $L_{N}$ into the equation and rearranging the terms algebraically gives:
\begin{equation*}
    \sum_{t=1}^{N} Len(c_{t}) + L_{N} = L_{0} - \sum_{t=1}^{N} p_\text{len}(c_{t})
\end{equation*}
which exactly matches Eq. \eqref{prop:length}.
\end{proof}

Eq. \eqref{prop:acc} reveals that the final accuracy of solving the problem is the baseline probability plus the sum of all step-wise accuracy increments ($\sum p_\text{acc}$). This means that as long as the model ensures $p_\text{acc} > 0$ at each step, it continuously strengthens the correct probability, preventing the model from falling into shallow reasoning or underthinking. Similarly, Eq. \eqref{prop:length} demonstrates that the total token cost of the current reasoning path equals the baseline expected length minus the accumulated length reductions ($\sum p_\text{len}$). In other words, every step with $p_\text{len} > 0$ acts as a ``shortcut'', bypassing redundant tokens and effectively suppressing verbosity and overthinking. Collectively, this property guarantees that guiding the model to generate steps with positive progress ($p_\text{acc} > 0$ and $p_\text{len} > 0$) is beneficial.

\subsection{Progress of Direct CoT and Vanilla CoT} \label{app:pro_example}
To further understand the concept of \textit{Progress}, we analyze the empirical progress of the specific reasoning trajectories provided in Example 2 (Vanilla CoT) and Example 3 (Direct CoT).

\textbf{The observation from Example 2} We observe obvious fluctuations in progress where several steps exhibit sharply negative progress values. These negative signals typically correspond to moments of model hesitation, redundant self-correction, or repetitive ``double-checking'' (e.g., \textit{``Wait, let me think again,''}). Notably, some self-correction steps increase the token consumption but fail to meaningfully improve the probability of generating the correct answer. This empirical observation aligns with recent literature \citep{wang2025wait}, which demonstrates that explicit self-reflection may degrade performance by generating redundant outputs rather than clarifying logic.

\begin{progressbox}[Example 2: Empirical Example of Progress in Vanilla CoT]

\textsc{Question:}\\[4pt]
Let $A$ equal the number of four digit odd numbers.  Let $B$ equal the number of four digit multiples of 5.  Find $A+B$.
\vspace{12pt}

\textsc{Progess in CoT:}\\[4pt]
\textit{\negprog{($p$ = [0.0, -618])} Alright, so I need to figure out A + B, where A is the number of four-digit odd numbers and B is the number of four-digit multiples of 5. Hmm, okay, let's break this down step by step.}
\textit{\posprog{($p$ = [+0.1, +1000])} First, let's tackle A, the number of four-digit odd numbers. A four-digit number ranges from 1000 to 9999. So, I need to count how many numbers in this range are odd. An odd number is one that ends with 1, 3, 5, 7, or 9. That gives me five possibilities for the last digit.} 
\textit{\negprog{($p$ = [-0.2, -108])} Now, to find the total number of four-digit numbers, it's 9000 because from 1000 to 9999, that's 9999 - 1000 + 1 = 9000 numbers. But since we're only considering odd numbers, I need to figure out how many of these 9000 are odd.} 
\textit{\negprog{($p$ = [+0.2, -1])} Since the last digit is what determines if a number is odd, I can calculate the number of four-digit numbers ending with 1, 3, 5, 7, or 9. Let me think: for the thousands place, it can't be zero, so it ranges from 1 to 9. That's 9 options. For the hundreds place, it can be any digit from 0 to 9, so that's 10 options. The tens place is also any digit from 0 to 9, another 10 options. And the units place, as we established, has 5 options (1,3,5,7,9).} 
\textit{\negprog{($p$ = [-0.1, +104])} So, the total number of four-digit odd numbers should be 9 (thousands) * 10 (hundreds) * 10 (tens) * 5 (units). Let me compute that: 9 * 10 is 90, 90 * 10 is 900, and 900 * 5 is 4500. So, A = 4500.} 
\textit{\negprog{($p$ = [+0.1, -1248])} Wait, that seems a lot. Let me check. From 1000 to 9999, half of them should be odd, right? Since every other number is odd, so 9000 / 2 = 4500. Yeah, that makes sense. So, A is indeed 4500.} 
\textit{\negprog{($p$ = [-0.2, +1835])} Now, moving on to B, the number of four-digit multiples of 5. A multiple of 5 ends with either 0 or 5. So, similar to before, let's figure out how many four-digit numbers end with 0 or 5.} 
\textit{\negprog{($p$ = [+0.2, -844])} Again, the thousands place is from 1 to 9 (9 options), hundreds place is 0-9 (10 options), tens place is 0-9 (10 options). The units place is 0 or 5, so 2 options.} 
\textit{\negprog{($p$ = [-0.1, +496])} Therefore, the total number of four-digit multiples of 5 is 9 * 10 * 10 * 2. Let's compute that: 9 * 10 is 90, 90 * 10 is 900, 900 * 2 is 1800. So, B = 1800.}
\textit{\posprog{($p$ = [0.0, +228])} Wait, let me think again. Since every fifth number is a multiple of 5, and four-digit numbers go from 1000 to 9999, how many multiples of 5 are there? Well, 9999 divided by 5 is approximately 1999.8, but since we're dealing with four-digit numbers, it's from 1000 to 9999 inclusive.} 
\textit{\negprog{($p$ = [0.0, -837])} Wait, the number of multiples of 5 in a range can be calculated by ((last - first)/step) + 1. So, ((9999 - 1000)/5) + 1. Let's compute that: (8999)/5 = 1799.8, which isn't an integer. Hmm, that might be a problem.} 
\textit{\posprog{($p$ = [0.0, +175])} Wait, no, actually, the number of multiples of 5 between 1000 and 9999 inclusive is equal to floor(9999/5) - floor(999/5). Let me compute floor(9999/5): 9999 /5 = 1999.8, so floor is 1999. floor(999/5) is floor(199.8) = 199. So, 1999 - 199 = 1800. So, that confirms B is 1800.} 
\textit{\posprog{($p$ = [+0.1, +534])} So, A is 4500 and B is 1800. Therefore, A + B is 4500 + 1800 = 6300.} 
\textit{\posprog{($p$ = [0.0, +1201])} Wait a second, just to make sure I didn't make any mistakes here. Let me think about another approach for B. Since the number must be a multiple of 5, it must end with 0 or 5. So, as I did earlier, thousands place 1-9 (9), hundreds 0-9 (10), tens 0-9 (10), units 0 or 5 (2). So, 91010*2=1800.} 
\textit{\posprog{($p$ = [0.0, +294])} And A is 4500, which is exactly half of 9000, which is the total number of four-digit numbers. So, that makes sense because every other number is odd, so half are odd.} 
\textit{\negprog{($p$ = [0.0, -397])} Therefore, adding them together, 4500 + 1800 equals 6300. So, A + B should be 6300.} 
\textit{\posprog{($p$ = [0.0, +503])} But wait, hold on, just to be thorough, let's double-check the number of four-digit numbers. 9999 - 1000 + 1 is indeed 9000, so that's correct. Half of 9000 is 4500, so A is 4500. For B, as we calculated, 1800. So, 4500 + 1800 is 6300.} 
\textit{\posprog{($p$ = [0.0, +276])} Yeah, that seems consistent. I think that's the answer. So, A + B is 6300.}
\textit{\posprog{($p$ = [0.0, +4])} Final Answer The value of $A + B$ is \boxed{6300}.}
\end{progressbox}

\textbf{The observation from Example 3} Conversely, the Direct CoT trajectory presents a different progress profile. Most of the progress values are positive, indicating a consistently advancement toward the solution. This demonstrates that extracting Direct CoT trajectories can yield a efficient reasoning step set. Interestingly, we also observe a phenomenon at the very first step of the Direct CoT sequence where the initial token triggers a massive progress gain ($p = [+0.0, +2530.7]$). This observation also aligns with the recent empirical findings of \citep{lincontrolling}. They observed that the specific leading words of the thought process (e.g., starting with ``To'' versus ``Okay'') dictate the verbosity of the entire subsequent output, demonstrating that forcing an efficient initial word substantially compresses the reasoning length. By extracting Direct CoT representations, our method implicitly leverages this powerful initial token to compress the output length.

\begin{progressbox}[Example 3: Empirical Example of Progress in Direct CoT]

\textsc{Question:}\\[4pt]
Let $A$ equal the number of four digit odd numbers.  Let $B$ equal the number of four digit multiples of 5.  Find $A+B$.
\vspace{12pt}

\textsc{Progess in CoT:}\\[4pt]
\textit{\posprog{($p$ = [+0.0, 2530.7])} To find $ A + B $, where $ A $ is the number of four-digit odd numbers and $ B $ is the number of four-digit multiples of 5, we proceed as follows:}
\textit{\posprog{($p$ = [+0.1, +83.3])}1. Calculating $ A $ (number of four-digit odd numbers): - A four-digit number ranges from 1000 to 9999. - The units digit must be an odd number (1, 3, 5, 7, 9), giving 5 choices. - The thousands digit can be any digit from 1 to 9, giving 9 choices. - The hundreds and tens digits can each be any digit from 0 to 9, giving 10 choices each. - Therefore, the total number of four-digit odd numbers is:$ A = 9 \times 10 \times 10 \times 5 = 4500 $}
\textit{\posprog{($p$ = [+0.0, +154])} 2. Calculating $ B $ (number of four-digit multiples of 5): - A number is a multiple of 5 if its units digit is 0 or 5, giving 2 choices. - The thousands digit can be any digit from 1 to 9, giving 9 choices. - The hundreds and tens digits can each be any digit from 0 to 9, giving 10 choices each. - Therefore, the total number of four-digit multiples of 5 is:$ B = 9 \times 10 \times 10 \times 2 = 1800 $} 
\textit{\posprog{($p$ = [+0.0, +30])} 3. Adding $ A $ and $ B $: - $ A + B = 4500 + 1800 = 6300 $} 
\textit{\posprog{($p$ = [0.0, +115])} Thus, the value of $ A + B $ is $\boxed{6300}$.}
\end{progressbox}

\section{More Analysis in Pilot Experiment}
\subsection{Visualization of Reasoning Steps under Different Models} \label{app:other_visual}

\begin{figure}[ht]
    \centering
    \includegraphics[width=\textwidth]{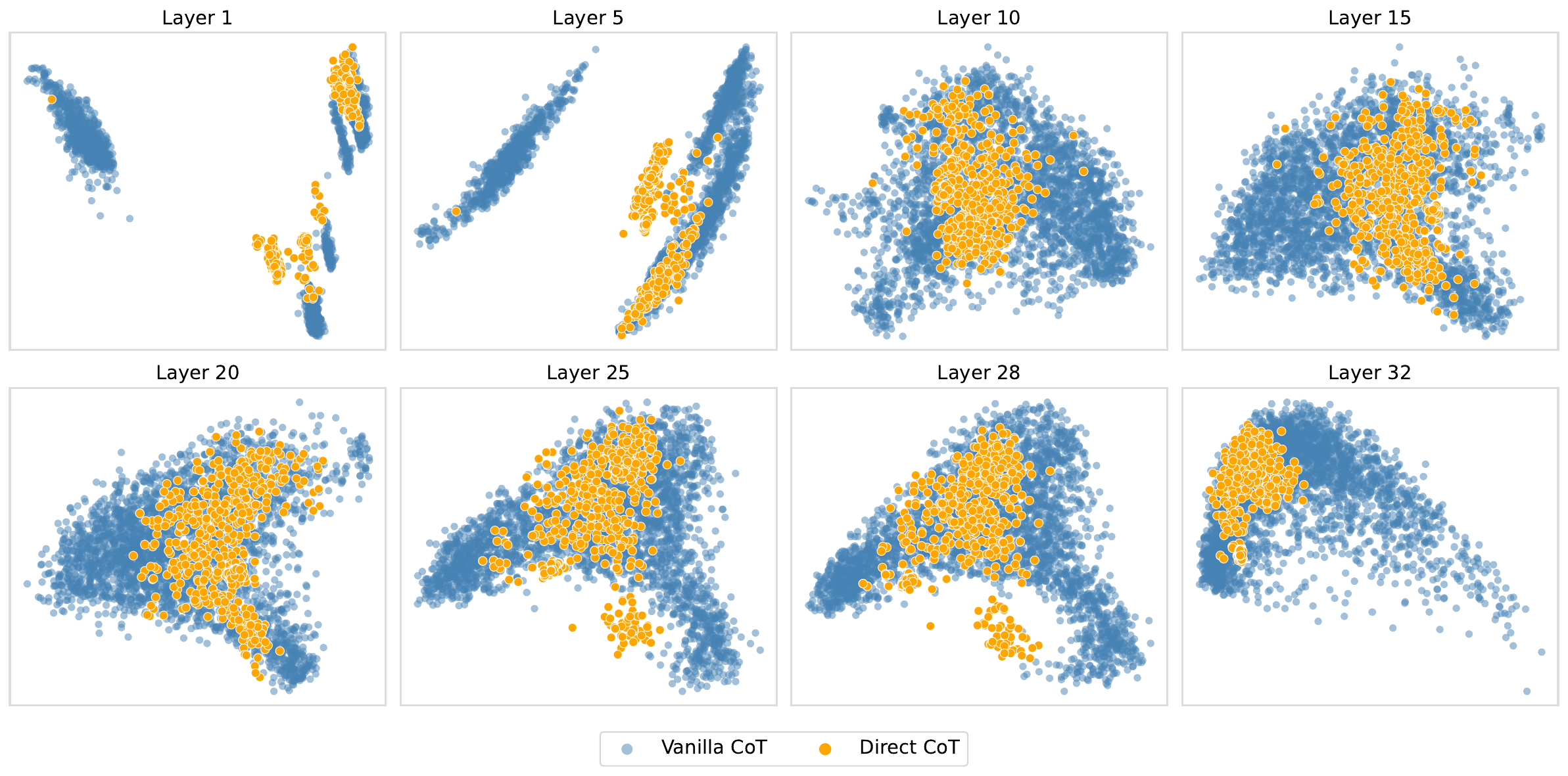} 
    \caption{Visualization of different reasoning steps generated by Qwen3-4B. The horizontal and vertical axes correspond to the first and second PCA components, respectively.}
    \label{fig:pca_4b}
\end{figure}

\begin{figure}[ht]
    \centering
    \includegraphics[width=\textwidth]{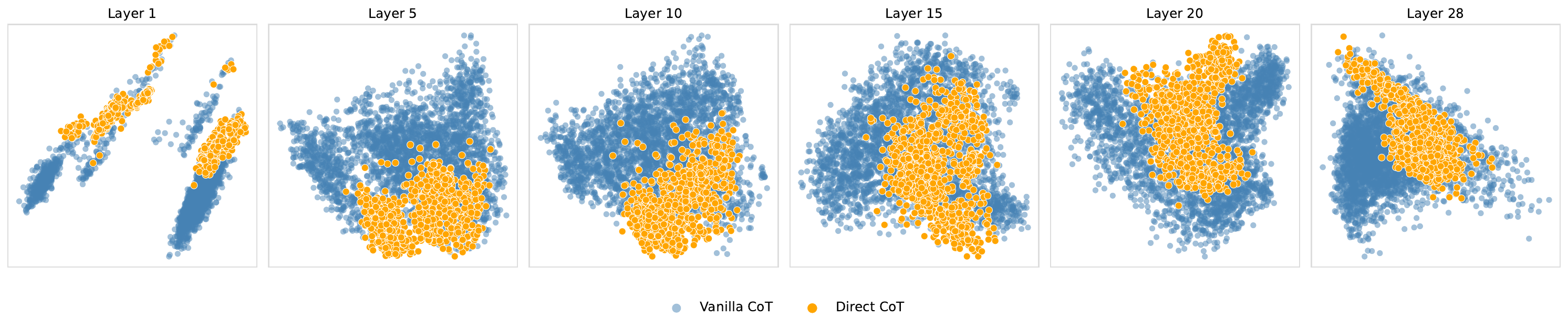} 
    \caption{Visualization of different reasoning steps generated by DeepSeek-R1-Distill-Qwen-7B. The horizontal and vertical axes correspond to the first and second PCA components, respectively.}
    \label{fig:pca_7b}
\end{figure}

\begin{figure}[ht]
    \centering
    \includegraphics[width=\textwidth]{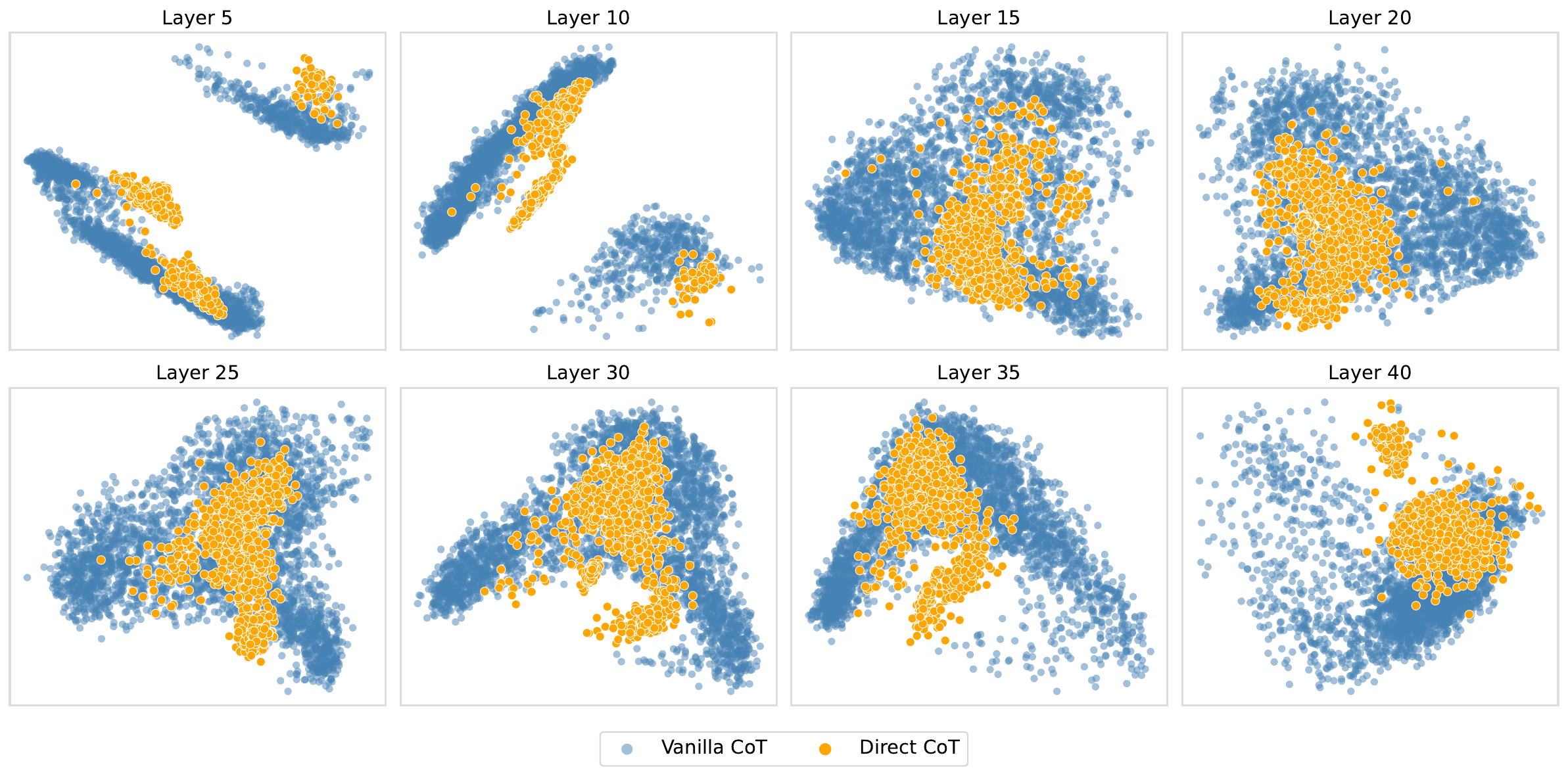} 
    \caption{Visualization of different reasoning steps generated by Qwen3-14B. The horizontal and vertical axes correspond to the first and second PCA components, respectively.}
    \label{fig:pca_14b}
\end{figure}

To verify the consistency of our geometric observation across different models, we extended the latent space visualization analysis described in Section \ref{sec:analysis} to DeepSeek-R1-Distill-Qwen-7B and Qwen3-4B/14B. As illustrated in Figure \ref{fig:pca_4b}, \ref{fig:pca_7b} and \ref{fig:pca_14b}, the experimental results across different models reveal a consistent phenomenon: the hidden representations of Direct CoTs naturally form a concentrated core region, whereas those of Vanilla CoTs are more dispersed and encapsulate this region. This reinforces our motivation to define this concentrated region as the proxy for the target set in Eq.~\eqref{obj:set}.

\subsection{The Further Analysis of Visualization in Latent Space}

\begin{figure}[ht]
    \centering
    \includegraphics[width=\textwidth]{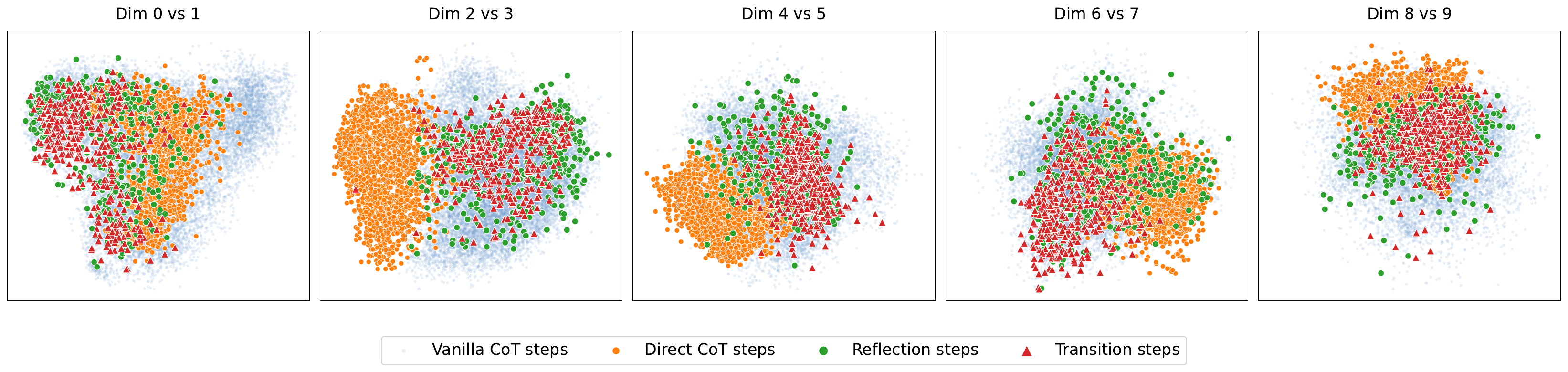} 
    \caption{Visualization of latent representations across different PCA dimensions. The results show that Direct CoT steps remain concentrated in a core region across various dimension pairs (e.g., Dim 0 vs 1, Dim 2 vs 3), while overthinking behaviors such as Transition and Reflection steps fall into the scattered region.}
    \label{fig:further_visual}
\end{figure}

We further explored the latent representations of the DeepSeek-R1-Distill-Qwen-1.5B model by examining higher-dimensional projections within the $k=10$ dimensional space. As illustrated in Figure \ref{fig:further_visual}, the concentration of Direct CoT steps relative to the scattered region of Vanilla CoT steps is not limited to the first two components but persists across multiple dimensions. This consistent geometric pattern reinforces our approach of adopting the core region formed by Direct CoT as an empirical proxy for the target set in Eq. (3).

To further understand how overthinking patterns deviate from this core region, we analyzed two specific behaviors typically observed in overthinking: Transition and Reflection steps, following the definitions in SEAL \citep{chenseal}. A Reflection step corresponds to a moment where the model pauses to verify its preceding logic, while a Transition step involves shifting the reasoning flow to explore a different perspective. Our analysis reveals that Direct CoT steps are clearly separable from these behaviors in the latent space. This clear separability illustrates that when hidden states deviate away from the efficient reasoning set, the model is likely engaging in unproductive hesitation or redundant self-correction.

\section{Additional Experimental Results} \label{app:add_exp}
\subsection{Cross-Domain and Cross-Difficulty Transferability} \label{app:cross}

To investigate the transferability of our method, we conduct experiments on DeepSeek-R1-Distill-Qwen-1.5B. Specifically, we construct efficient reasoning sets by sampling Direct CoTs and Vanilla CoTs from various datasets, encompassing different mathematical difficulty levels (GSM8K, MATH500, AMC 2023, AIME 2025) as well as a distinct scientific domain (GPQA). For GSM8K and MATH500, we randomly select 300 questions and generate one Direct CoT and one Vanilla CoT per question. For GPQA, we generate one Direct CoT and one Vanilla CoT for each question in the dataset. For AMC 2023 and AIME 2025, due to the limited number of available questions, we oversample by generating 10 CoT per question to reach a comparable scale.

As shown in Table \ref{tab:combined_performance}, applying efficient reasoning sets constructing from different datasets consistently outperforms the Vanilla baseline in both accuracy and token reduction across all evaluation benchmarks. Remarkably, even when the efficient reasoning set is constructed from easy math (GSM8K) or scientific questions (GPQA), it still effectively guides the model to solve highly complex competition-level math (AIME 2025) and coding tasks (LiveCodeBench). This transferability demonstrates that the efficient reasoning set $\mathcal{H}$ captures a general, domain-agnostic geometric property of efficient reasoning in the latent space.

Furthermore, we observe that the efficient reasoning set constructed from the MATH500 dataset yields a highly effective and balanced performance across both in-domain and out-of-domain evaluations. Consequently, we adopt the MATH500-derived set as our default experimental setting throughout the main text.

\begin{table}[htbp]
\centering
\caption{Cross-Domain and Cross-Difficulty Transferability on DeepSeek-R1-Distill-Qwen-1.5B.}
\label{tab:combined_performance}
\small
\renewcommand{\arraystretch}{1.2}
\setlength{\tabcolsep}{2.5pt}
\resizebox{\textwidth}{!}{
\begin{tabular}{lcccccccccccc}
\toprule
\multirow{2}{*}{\textbf{Methods}} & \multicolumn{2}{c}{GSM8K} & \multicolumn{2}{c}{MATH500} & \multicolumn{2}{c}{AMC2023} & \multicolumn{2}{c}{AIME2025} & \multicolumn{2}{c}{LiveCodeBench} & \multicolumn{2}{c}{GPQA-Diamond} \\
\cmidrule(lr){2-3} \cmidrule(lr){4-5} \cmidrule(lr){6-7} \cmidrule(lr){8-9} \cmidrule(lr){10-11} \cmidrule(lr){12-13}
 & Acc. & Tok. & Acc. & Tok. & Acc. & Tok. & Acc. & Tok. & Acc. & Tok. & Acc. & Tok. \\
\midrule
Vanilla & 81.5 & 1489 & 82.1 & 4476 & 64.4 & 7708 & 17.3 & 11811 & 32.1 & 10013 & 33.3 & 8241 \\
Ours-MATH500 & 83.1 & 853 & 83.2 & 2949 & 75.2 & 4821 & 26.7 & 8502 & 32.7 & 7656 & 36.4 & 5651 \\
Ours-GSM8K & 82.3 & 707 & 82.3 & 2980 & 76.7 & 4825 & 23.3 & 8097 & 32.8 & 7575 & 35.8 & 5738 \\
Ours-AMC23 & 82.6 & 782 & 82.9 & 2951 & 72.5 & 4609 & 22.2 & 8649 & 32.5 & 7594 & 35.3 & 5590 \\
Ours-AIME2025 & 82.7 & 901 & 82.0 & 3294 & 71.6 & 4939 & 30.0 & 8727 & 33.4 & 8393 & 34.9 & 6149 \\
Ours-GPQA & 81.6 & 843 & 82.4 & 3141 & 70.0 & 5377 & 25.5 & 9145 & 32.2 & 8411 & 34.8 & 5827 \\
\bottomrule
\end{tabular}
}
\end{table}

\subsection{Analysis of Semantic Change} \label{app:semantic}
To further analyze the semantic changes of the model's output after our method's intervention, we conduct a statistical semantic analysis over all generated reasoning trajectories on the MATH500 dataset. Following the vocabularies introduced in SEAL \citep{chenseal}, we explicitly quantify the occurrence of Transition and Reflection vocabularies by reporting their average word count, term frequency (TF), and term frequency-inverse document frequency (TF-IDF) scores.. For readability, the TF and TF-IDF scores are scaled by a factor of $1,000$. These metrics are evaluated alongside the final reasoning Accuracy and overall Token consumption.

As detailed in Table \ref{tab:semantic_analysis}, we observe a consistent reduction in the frequency of these specific reasoning steps across all evaluated models. This indicates that our method effectively limits the model's tendency to engage in extensive self-verification (Reflection) and perspective shifts (Transition). Notably, this reduction directly correlates with lower overall token consumption without compromising reasoning accuracy. 

\begin{table}[htbp]
\centering
\caption{Semantic statistics of Transition and Reflection vocabularies.}
\label{tab:semantic_analysis}
\small
\renewcommand{\arraystretch}{1.1}
\begin{tabular}{l ccc ccc cc}
\toprule
\multirow{2}{*}{Method} & \multicolumn{3}{c}{Reflection} & \multicolumn{3}{c}{Transition} & \multicolumn{2}{c}{Performance} \\
\cmidrule(lr){2-4} \cmidrule(lr){5-7} \cmidrule(lr){8-9}
 & Word Count & TF & TF--IDF & Word Count & TF & TF--IDF & Acc. & Tok. \\
\midrule
\multicolumn{9}{l}{\textbf{DeepSeek-R1-Distill-Qwen-1.5B}} \\
Baseline & 28.5 & 9.3 & 13.0 & 8.0 & 2.5 & 4.7 & 82.1 & 4476 \\
Ours     & 6.0 & 4.2 & 6.6 & 1.7 & 0.7 & 1.5 & 83.2 & 2949 \\
\midrule
\multicolumn{9}{l}{\textbf{DeepSeek-R1-Distill-Qwen-7B}} \\
Baseline & 18.3 & 7.6 & 10.3 & 4.9 & 2.1 & 3.7 & 91.6 & 3647 \\
Ours     & 6.1 & 4.3 & 6.4 & 1.9 & 1.0 & 1.7 & 92.2 & 2599 \\
\midrule
\multicolumn{9}{l}{\textbf{Qwen3-4B}} \\
Baseline & 32.2 & 9.0 & 11.5 & 11.5 & 4.0 & 5.5 & 95.2 & 4587 \\
Ours     & 15.3 & 6.7 & 8.9 & 3.5 & 1.5 & 2.3 & 95.8 & 3550 \\
\midrule
\multicolumn{9}{l}{\textbf{Qwen3-14B}} \\
Baseline & 22.3 & 7.8 & 12.5 & 7.3 & 2.8 & 4.9 & 96.2 & 4559 \\
Ours     & 3.6 & 2.0 & 3.4 & 1.0 & 0.6 & 1.1 & 96.8 & 3510 \\
\bottomrule
\end{tabular}
\end{table}

\subsection{Creativity Analysis} \label{app:creativity} 
While our method is specifically developed to improve reasoning capacity, a natural question arises: could this method introduce unintended side effects on other fundamental capabilities? To address this concern, we extend our analysis to investigate the method's influence on other general model abilities. Specifically, we focus on exploring whether it negatively impacts creativity or the naturalness of linguistic expressions. To conduct this evaluation, we assess the models using the Creative Writing v3 benchmark \citep{creative-writing-bench-v3}, which is also adopted in recent studies \citep{liefficient, qwen3technicalreport}. The detailed results of this assessment are presented in Table \ref{tab:creative_writing}. 

In our evaluation, we utilize officially recommended Claude-Sonnet-4.0 as the judge model. Specifically, we focus on two primary metrics: 
(1) \textbf{Rubric Score}, which provides a comprehensive evaluation of overall writing quality across various dimensions; 
(2) \textbf{Fine-grained Ability Scores}, which measure performance across fifteen distinct creative writing skills. These competencies include \textit{coherence} (logical consistency and clear structure), \textit{creativity} (originality and avoidance of templates), \textit{descriptive imagery} (vivid and sensory language), \textit{pacing} (appropriate narrative flow), \textit{elegant prose} (stylistic fluency), \textit{instruction following} (strict adherence to prompts), \textit{consistent voice and tone} (stable narrative perspective), \textit{strong dialogue} (natural and character-driven conversations), \textit{sentence flow} (smooth transitions), \textit{show-don't-tell} (conveying meaning through actions rather than direct exposition), \textit{avoidance of amateurish prose} (steering clear of clichéd patterns), \textit{emotional depth} (nuanced feelings), \textit{avoidance of positivity bias} (rejecting forced optimism), \textit{avoidance of purple prose} (restraint from overly ornate vocabulary), and \textit{believable characters} (psychological realism).

As demonstrated in the results, applying our method does not compromise the semantic richness of the outputs. Instead, the models largely preserve, and in several cases even enhance, their performance regarding creative expression and natural language fluency.

\begin{table*}[t]
\centering
\caption{Creative-writing performance on Creative Writing v3 benchmark. Ability abbreviations: Coh = Coherent; Crt = Creativity; Img = Descriptive Imagery; Pac = Pacing; Ele = Elegant Prose; Inst = Instruction Following; Voi = Consistent Voice \& Tone; Dia = Strong Dialogue; Flo = Sentence Flow; SDT = Show-Don’tTell; Ama = Avoids Amateurish Prose; Emo = Emotional Depth; Pos = Avoids Positivity Bias; Pur = Avoids Purple Prose; Ch = Believable Characters.}
\label{tab:creative_writing}
\resizebox{\textwidth}{!}{
\begin{tabular}{lcccccccccccccccc}
\toprule
\multirow{2}{*}{Method} & \multicolumn{1}{c}{Score} & \multicolumn{15}{c}{Ability} \\
\cmidrule(lr){2-2} \cmidrule(lr){3-17}
& Rubric & Coh & Crt & Img & Pac & Ele & Inst & Voi & Dia & Flo & SDT & Ama & Emo & Pos & Pur & Ch \\
\midrule
\multicolumn{17}{l}{\textbf{DeepSeek-R1-Distill-Qwen-1.5B}} \\
Baseline & 13.80 & 0.55 & 1.20 & 1.09 & 1.20 & 0.30 & 0.39 & 0.84 & 1.33 & 0.48 & 1.38 & 0.91 & 0.19 & \textbf{16.88} & \textbf{14.83} & 0.28 \\
Ours & \textbf{14.16} & \textbf{0.75} & \textbf{1.53} & \textbf{1.27} & \textbf{1.42} & \textbf{0.45} & \textbf{0.50} & \textbf{0.93} & \textbf{1.37} & \textbf{0.60} & \textbf{1.61} & \textbf{1.33} & \textbf{0.31} & 14.60 & 13.78 & \textbf{0.52} \\
\midrule
\multicolumn{17}{l}{\textbf{DeepSeek-R1-Distill-Qwen-7B}} \\
Baseline & 21.40 & 3.60 & 3.20 & 3.92 & 4.98 & 2.30 & \textbf{2.11} & 3.89 & 3.36 & 3.01 & 3.83 & 2.78 & \textbf{1.73} & \textbf{12.48} & \textbf{11.72} & 2.84 \\
Ours & \textbf{22.56} & \textbf{4.19} & \textbf{3.99} & \textbf{4.07} & \textbf{5.81} & \textbf{2.40} & 2.10 & \textbf{4.21} & \textbf{4.26} & \textbf{3.31} & \textbf{4.24} & \textbf{3.57} & 1.62 & 12.22 & 11.62 & \textbf{3.15} \\
\midrule
\multicolumn{17}{l}{\textbf{Qwen3-4B}} \\
Baseline & 36.83 & 10.98 & \textbf{4.71} & \textbf{7.89} & \textbf{10.56} & 5.89 & 5.35 & 10.02 & 5.02 & 8.29 & \textbf{4.73} & \textbf{4.93} & 3.79 & \textbf{13.56} & 11.61 & 7.88 \\
Ours & \textbf{37.44} & \textbf{12.05} & 4.64 & 7.75 & 8.85 & \textbf{6.12} & \textbf{5.56} & \textbf{10.62} & \textbf{5.18} & \textbf{8.62} & 4.45 & 4.92 & \textbf{4.11} & 13.36 & \textbf{12.25} & \textbf{8.22} \\
\midrule
\multicolumn{17}{l}{\textbf{Qwen3-14B}} \\
Baseline & 59.06 & 16.41 & \textbf{6.32} & 12.45 & \textbf{15.45} & 10.74 & 11.74 & 15.21 & 8.90 & 13.56 & \textbf{8.44} & 8.67 & 8.30 & \textbf{15.83} & \textbf{14.37} & 13.74 \\
Ours & \textbf{59.89} & \textbf{16.73} & 6.18 & \textbf{12.47} & 14.91 & \textbf{10.86} & \textbf{13.37} & \textbf{15.43} & \textbf{9.79} & \textbf{13.60} & 8.40 & \textbf{9.09} & \textbf{8.79} & 14.86 & 14.07 & \textbf{13.96} \\
\bottomrule
\end{tabular}
}
\end{table*}

\subsection{Pass@k Performance Analysis} \label{app:pass_k}
To further analyze whether our method affects the model's exploration capability, we conduct a Pass@$k$ evaluation. Pass@$k$ measures the probability that at least one out of $k$ independently generated reasoning trajectories yields the correct answer. It is widely used to assess the exploration capability of a model's generation distribution \citep{karan2025reasoning}. Specifically, we evaluate our method on two hard mathematical benchmarks, AMC2023 and AIME2025. We report the Pass@1, Pass@4, Pass@8, and Pass@16 scores in Table \ref{tab:pass_k_performance}. 
The results show that our method consistently maintains, and even improves, the Pass@$k$ performance across various model scales. For example, on the AIME 2025 dataset using the DeepSeek-R1-Distill-Qwen-7B model, our method improves Pass@1 from 26.9\% to 39.0\% and Pass@16 from 66.7\% to 70.0\%. The results demonstrate that projecting hidden states into the efficient reasoning set $\mathcal{H}$ does not collapse the model's output diversity nor harm its ability to navigate complex problem spaces. Instead, by pruning unproductive deviations, the model concentrates its exploration capacity on highly promising reasoning trajectories, achieving a better balance between reasoning diversity and generation efficiency.

\begin{table}[htbp]
\centering
\caption{Pass@k Performance on AMC 2023 and AIME 2025.}
\label{tab:pass_k_performance}
\small
\renewcommand{\arraystretch}{1.2}
\setlength{\tabcolsep}{2.5pt}
\resizebox{0.8\textwidth}{!}{
\begin{tabular}{lcccccccc}
\toprule
\multirow{2}{*}{\textbf{Methods}} & \multicolumn{4}{c}{AMC 2023} & \multicolumn{4}{c}{AIME 2025}  \\
\cmidrule(lr){2-5} \cmidrule(lr){6-9} 
 & Pass@1 & Pass@4 & Pass@8 & Pass@16 & Pass@1 & Pass@4 & Pass@8 & Pass@16 \\
\midrule
\multicolumn{9}{l}{\textbf{DeepSeek-R1-Distill-Qwen-1.5B}} \\
Vanilla & 64.4 & 85.6 & 92.5 & 95.0  & 17.3 & 33.3 & 40.0 & 40.0 \\
Ours    & \textbf{75.2} & \textbf{91.2} & \textbf{96.2} & \textbf{97.5} & \textbf{26.7} & \textbf{36.7} & \textbf{40.0} & \textbf{40.0} \\
\midrule
\multicolumn{9}{l}{\textbf{DeepSeek-R1-Distill-Qwen-7B}} \\
Vanilla & 86.3 & 94.3 & 95.0 & 95.0 & 26.9 & 58.3 & 63.3 & 66.7 \\
Ours    & \textbf{91.1} & \textbf{96.3} & \textbf{97.5} & \textbf{100.0} & \textbf{39.0} & \textbf{60.0} & \textbf{66.7} & \textbf{70.0}  \\
\midrule
\multicolumn{9}{l}{\textbf{Qwen3-4B}} \\
Vanilla & 95.9 & 100.0 & 100.0 & 100.0 & 60.2 & 70.0 & 81.7 & 86.7 \\
Ours    & \textbf{97.5} & \textbf{100.0} & \textbf{100.0} & \textbf{100.0} & \textbf{67.1} & \textbf{78.3} & \textbf{83.3} & \textbf{86.7} \\
\midrule
\multicolumn{9}{l}{\textbf{Qwen3-14B}} \\
Vanilla & 96.9 & 100.0 & 100.0 & 100.0 & 67.1 & 76.7 & 83.3 & 83.3 \\
Ours    & \textbf{98.3} & \textbf{100.0} & \textbf{100.0} & \textbf{100.0} & \textbf{70.4} & \textbf{76.7} & \textbf{83.3} & \textbf{83.3} \\
\bottomrule
\end{tabular}
}
\end{table}

\subsection{Statistical Reliability and Significance Testing}
\label{app:statistical_reliability}
For small-sized benchmark datasets such as AMC2023 and AIME2025, we report the mean accuracy (\%), average token consumption per problem, and corresponding standard deviations to establish statistical reliability. Across all evaluated models, our approach yields statistically significant accuracy improvements alongside substantial reductions in token consumption.

\begin{table}[h!]
\centering
\small
\caption{\textbf{Statistical reliability evaluation on AMC23 and AIME2025.} We report mean accuracy, mean token count and corresponding standard deviations across datasets and model sizes.}
\label{tab:statistical_reliability}
\begin{tabular}{l l c c c c}
\toprule
\multirow{2}{*}{\textbf{Model}} & \multirow{2}{*}{\textbf{Method}} & \multicolumn{2}{c}{\textbf{AMC2023}} & \multicolumn{2}{c}{\textbf{AIME2025}} \\
\cmidrule(lr){3-4} \cmidrule(lr){5-6}
 & & \textbf{Acc} & \textbf{Tokens} & \textbf{Acc} & \textbf{Tokens} \\
\midrule
\multirow{2}{*}{DeepSeek-1.5B} 
 & Vanilla & 64.4 $\pm$ 5.4 & 7708 $\pm$ 703 & 17.3 $\pm$ 3.7 & 11811 $\pm$ 625 \\
 & Ours    & 75.2 $\pm$ 3.5 & 4821 $\pm$ 304 & 26.7 $\pm$ 4.2 & 8502 $\pm$ 390 \\
\midrule
\multirow{2}{*}{DeepSeek-7B} 
 & Vanilla & 86.3 $\pm$ 4.0 & 5846 $\pm$ 358 & 26.9 $\pm$ 3.9 & 11209 $\pm$ 780 \\
 & Ours    & 91.1 $\pm$ 2.7 & 4083 $\pm$ 261 & 39.0 $\pm$ 4.0 & 9042 $\pm$ 631 \\
\midrule
\multirow{2}{*}{Qwen-4B} 
 & Vanilla & 95.9 $\pm$ 1.7 & 7588 $\pm$ 475 & 60.2 $\pm$ 3.7 & 16835 $\pm$ 944 \\
 & Ours & 97.5 $\pm$ 1.3 & 5898 $\pm$ 187 & 67.1 $\pm$ 5.2 & 14809 $\pm$ 525 \\
\midrule
\multirow{2}{*}{Qwen3-14B} 
 & Vanilla & 96.9 $\pm$ 1.5 & 7277 $\pm$ 336 & 67.1 $\pm$ 3.7 & 16019 $\pm$ 888 \\
 & Ours    & 98.3 $\pm$ 1.3 & 5751 $\pm$ 296 & 70.4 $\pm$ 3.8 & 13144 $\pm$ 603 \\
\bottomrule
\end{tabular}
\end{table}

\section{Details on Solving Quadratic Programming}\label{app:details_on_qp}

In this section, we provide a comprehensive discussion on how we formulate and efficiently solve the projection problem \eqref{eq:projection}.

\subsection{The Definition of Quadratic Programming}
A standard Quadratic Programming (QP) problem aims to minimize a quadratic objective function subject to linear constraints. The general formulation is given by:
\begin{equation}
    \min_{u \in \mathbb{R}^k} \frac{1}{2} u^\top P u + q^\top u \quad \text{s.t.} \quad G u \le h
\end{equation}
where $P \in \mathbb{R}^{k \times k}$ is a symmetric positive-definite matrix, $q \in \mathbb{R}^k$ is a linear coefficient vector, $G \in \mathbb{R}^{M \times k}$ represents the constraint matrix, and $h \in \mathbb{R}^M$ is the constraint boundary vector. 

In our method, the projection operation seeks the closest point $z^*_t$ within the efficient reasoning set $\mathcal{H}$ by minimizing its Euclidean distance to the original hidden state $z_t$:
\begin{equation} 
    z^*_t = \arg\min_{u \in \mathbb{R}^k} \frac{1}{2} \|z_t - u\|_2^2 \quad \text{s.t.} \quad w_m^\top u + b'_m \le 0, \forall \ m \in \{1, \dots, M\}
\end{equation}
By expanding the objective function $\frac{1}{2} \|z_t - u\|_2^2 = \frac{1}{2} u^\top u - z_t^\top u + \frac{1}{2} z_t^\top z_t$, and omitting the constant term $\frac{1}{2} z_t^\top z_t$ which does not affect the optimization, we can exactly map our problem to the standard QP format:
\begin{itemize}
    \item The quadratic matrix $P$ is the identity matrix $I \in \mathbb{R}^{k \times k}$.
    \item The linear coefficient vector $q$ is $-z_t$.
    \item The constraint matrix $G \in \mathbb{R}^{M \times k}$ is composed of the normal vectors $w_m^\top$.
    \item The constraint boundary vector $h \in \mathbb{R}^M$ is composed of the elements $-b'_m$.
\end{itemize}

Crucially, in our method, the constraint boundaries defining $\mathcal{H}$ (i.e., $G$ and $h$) are pre-computed and fixed during inference. However, the linear term $q = -z_t$ varies dynamically at each reasoning step. Because the optimization problem is parameterized by the dynamically changing input vector $z_t$, it is also classified as a Multi-Parametric Quadratic Programming (mp-QP) problem.

\subsection{Online Numerical Solving Approach}
For a given input state $z_t$, the mp-QP is instantiated into a specific standard QP instance. Typically, solving such instances relies on iterative numerical optimization algorithms, such as the active-set method \citep{arnstrom2022dual} and first-order method \citep{stellato2020osqp}. These numerical methods are highly robust and have been integrated into well-established solvers.

For online solving, we utilize \texttt{CvxpyLayer}, a differentiable convex optimization solver, to compute the optimal solution $z^*_t$ during the model's forward pass. While numerically reliable, this online approach suffers from a major bottleneck: numerical solvers typically execute on the CPU. Consequently, for every intervention, the inference engine must transfer the hidden state $z_t$ from the GPU to the CPU, wait for the solver to converge, and transfer $z^*_t$ back. This frequent CPU-GPU communication introduces additional latency for high-throughput LLM serving.

\subsection{Explicit Acceleration via Offline mp-QP}
To eliminate the CPU-GPU communication bottleneck, we exploit a fundamental structural property of mp-QP: the optimal solution $z^*_t$ can be analytically expressed as a continuous piecewise affine function of the input parameter $z_t$ \citep{bemporad2002explicit}.

Specifically, the continuous $k$-dimensional latent space can be partitioned into a finite set of non-overlapping critical regions $\{\mathcal{CR}_1, \dots, \mathcal{CR}_N\}$. Within each specific region $\mathcal{CR}_i$, the optimal solution $z^*_t$ is strictly given by a closed-form affine mapping:
\begin{equation}
    z^*_t = F_i z_t + g_i, \quad \text{if } z_t \in \mathcal{CR}_i
\end{equation}
where $F_i \in \mathbb{R}^{k \times k}$ and $g_i \in \mathbb{R}^k$ are pre-computable matrices. Each critical region $\mathcal{CR}_i$ is a convex polyhedral set characterized by linear inequalities, i.e., $\mathcal{CR}_i = \{z \in \mathbb{R}^k \mid H_i z \le K_i\}$.

We leverage the \texttt{PDAQP} solver to perform this partitioning and matrix calculation offline \citep{arnstrom2024pdaqp}. During online deployment, solving the complex optimization problem is decomposed into two pure GPU tensor operations: (1) Region Querying, which identifies the index $i$ such that $z_t \in \mathcal{CR}_i$; and (2) Affine Transformation, which directly applies $F_i z_t + g_i$ to obtain the solution. This computation operates entirely on the GPU, completely bypassing the CPU and ensuring high-throughput reasoning.

\subsection{Computational Overhead Analysis} \label{app:overhead}
The explicit acceleration above heavily relies on the scale of the problem. Because our method intervenes in a highly compressed latent space, the QP involves only a very low dimension of 10 to 16 decision variables and exactly $M=8$ linear constraints. This prevents the number of critical regions from combinatorial explosion, making offline calculation and storage extremely lightweight.

\paragraph{Experimental Design.} To evaluate the computational overhead presented in Table \ref{tab:solving_time}, we collected 1,000 instances of the parameter $z_t$ generated during the actual inference process across different models. We compared the online QP solving method (denoted as \textit{Online}) with our accelerated approach (denoted as \textit{Explicit}). Specifically, the online method directly invokes the numerical solver CvxpyLayer for each QP instance during the model's forward pass. The explicit method relies on offline mp-QP precomputation executed using the PDAQP solver \citep{arnstrom2024pdaqp}, followed by online region querying implemented via PyTorch operators. The \textit{Additional Memory} metric quantifies the GPU memory footprint required to store the pre-computed piecewise linear expressions for the explicit method.

\paragraph{Results and Conclusions.} As demonstrated in Table \ref{tab:solving_time}, the explicit acceleration yields dramatic efficiency gains. The online solving method (CvxpyLayer) requires between 1.7 ms to 5.3 ms per operation across different models. In contrast, our explicit GPU-accelerated approach reduces this latency to a mere 0.3 ms to 0.5 ms, achieving approximately a $10\times$ speedup. Furthermore, storing the offline-computed piecewise linear expressions (i.e., matrices $F_i$ and $g_i$) introduces an exceptionally small GPU memory footprint, ranging from just 0.41 MB for the 1.5B model to 1.80 MB for the 14B model. This negligible memory cost ensures that our acceleration framework can be seamlessly integrated into environments with resource constraints.

\begin{table}[htbp]
\centering
\caption{Comparison of Average Solving Time. We detail the scale of the QP problem and compare the latency of the online solver with the explicit acceleration. The additional memory denotes the GPU footprint required to store the precomputed piecewise linear mappings for the explicit method. }
\label{tab:solving_time}
\small
\begin{tabular}{lcccccc}
\toprule
\textbf{Model} & \textbf{Variables} & \textbf{Constraints} & \textbf{Online (ms)} & \textbf{Explicit (ms)} & \textbf{Additional Memory (MB)} \\
\midrule
R1-1.5B  & 10 & 8 & 1.7 & 0.3 & 0.41 \\
R1-7B    & 12 & 8 & 2.3 & 0.3 & 0.93 \\
Qwen3-4B & 12 & 8 & 2.9 & 0.3 & 0.56 \\
Qwen3-14B & 16 & 8 & 5.3 & 0.5 & 1.80 \\
\bottomrule
\end{tabular}
\end{table}

\section{Details on Experiment Settings} \label{app:experment_setting}
In this section, we provide a comprehensive overview of our experimental setup, including the mathematical notations used throughout the paper, the specific prompt templates for generating reasoning trajectories, and the detailed hyperparameter settings and decoding configurations for the models evaluated.

\subsection{Notation Summary}
To facilitate reading and ensure mathematical clarity, we summarize the key notations used in our problem formulation, latent space analysis, and method in Table \ref{tab:notation_summary}.

\begin{table}[htbp]
    \centering
    \caption{Mathematical notations and their descriptions used in our method.}
    \label{tab:notation_summary}
    \renewcommand{\arraystretch}{1.1} 
    \begin{tabular}{c | p{11.5cm}}
        \toprule
        \textbf{Symbol} & \textbf{Description} \\
        \midrule
        $X_{dir}, X_{van}$ & The original representations of Direct CoTs and Vanilla CoTs. \\
        $Z_{dir}, Z_{van}$ & The representations of Direct CoTs and Vanilla CoTs after PCA projection. \\
        $\mathcal{H}$ & The efficient reasoning. \\
        $U$ & The PCA projection matrix. \\
        $k$ & The dimension of the latent space after the PCA projection. \\
        $M$ & The number of clusters defining the efficient reasoning set $\mathcal{H}$. \\
        $\alpha$ & The boundary shift proportion, determining the inclusivity of the set $\mathcal{H}$  \\
        $\lambda$ & The intervention strength controlling the magnitude of the correction. \\
        \bottomrule
    \end{tabular}
\end{table}

\subsection{Prompt Structure} \label{app:direct}

We first prompt the model using the ground-truth answer along with conciseness instructions. During generation, we discard the intermediate \texttt{<think>} block and keep only the finalized deduction generated after the \texttt{</think>} token. Note that for some problems exceeding the model's reasoning capability, the model may fail to produce a valid deduction and hits the maximum response length limit. We filter out these incomplete generations. 

To extract hidden representations, we concatenate this finalized deduction with the initial question prompt (i.e. Vanilla CoT prompt) to construct a synthetic sequence $[\text{Vanilla CoT Prompt}] + \texttt{<think>} + [\text{Direct CoT}]$. Then we perform a forward pass on this synthetic sequence and extract the hidden representations at step delimiters (e.g., \texttt{\textbackslash n\textbackslash n}). The prompt templates used in our method are provided below.

\begin{tcolorbox}[
    enhanced, 
    colframe=blue!50!gray,
    colback=blue!3,
    colbacktitle=blue!10,
    coltitle=blue!40!black,
    boxrule=0.8pt,
    arc=3pt,
    boxsep=0pt,
    left=2mm,
    right=2mm,
    top=3mm,
    bottom=3mm,
    toptitle=3mm,
    bottomtitle=3mm,
    title=\textbf{The Prompt of Vanilla CoT},
    fonttitle=\normalsize\sffamily,
    fontupper=\normalsize,
    before upper={\setlength{\parskip}{6pt}}, 
    before upper={\setlength{\parindent}{0pt}}, 
    parbox=false 
]

Answer the following questions. You should think step-by-step and put your final answer within \textbackslash \textbackslash boxed\{\}.

Question: \{QUESTION\}

\end{tcolorbox}

\begin{tcolorbox}[
    enhanced, 
    colframe=blue!50!gray,
    colback=blue!3,
    colbacktitle=blue!10,
    coltitle=blue!40!black,
    boxrule=0.8pt,
    arc=3pt,
    boxsep=0pt,
    left=2mm,
    right=2mm,
    top=3mm,
    bottom=3mm,
    toptitle=3mm,
    bottomtitle=3mm,
    title=\textbf{The Prompt for Generating Direct CoT},
    fonttitle=\normalsize\sffamily,
    fontupper=\normalsize,
    before upper={\setlength{\parskip}{6pt}}, 
    before upper={\setlength{\parindent}{0pt}}, 
    parbox=false 
]

Given a question and its answer, provide a detailed derivation.

\#\#\# Requirements

1. DO NOT use headers like "Step-by-Step" or other subtitles(e.g. "Step 1").

2. ONLY provide the sequential logic and the necessary math steps in a cohesive flow.

3. Ensure the derivation is compact and direct.

4. The result must conclude with the final answer in \textbackslash \textbackslash boxed\{\}.

\#\#\# Task

Question: \{QUESTION\}

Answer: \{ANSWER\}

\end{tcolorbox}

\subsection{Hyperparameter and Decoding Settings.}
\textbf{Decoding Settings.} Following the officially recommended configurations for each model, we utilize a top-$p$ of 0.95, a temperature of 0.6. The maximum sequence length is set to 16,384 tokens for the DeepSeek-R1-Distill-Qwen-1.5B/7B models, and 32,768 tokens for the Qwen3-4B/14B model.

\textbf{Hyperparameter Settings.} Based on the extensive ablation studies detailed in Appendix \ref{app:ablation}, we use a fixed intervention strength $\lambda = 1.0$, cluster number $M = 8$, and boundary proportion $\alpha = 90\%$ across all models. The steering layer and PCA dimension $k$ are configured as follows: DeepSeek-R1-Distill-Qwen-1.5B (layer 20, $k=10$), DeepSeek-R1-Distill-Qwen-7B (layer 20, $k=12$), Qwen3-4B (layer 25, $k=12$), and Qwen3-14B (layer 30, $k=16$).

\section{Details on Ablation Study} \label{app:ablation}
In this section, we present the comprehensive results of tuning key hyperparameters across three models: DeepSeek-R1-Distill-Qwen-1.5B, DeepSeek-R1-Distill-Qwen-7B, and Qwen3-4B on the MATH500 benchmark.

\subsection{Impact of Intervention Strength $\lambda$}

\begin{figure}[htbp]
    \centering
    \includegraphics[width=\textwidth]{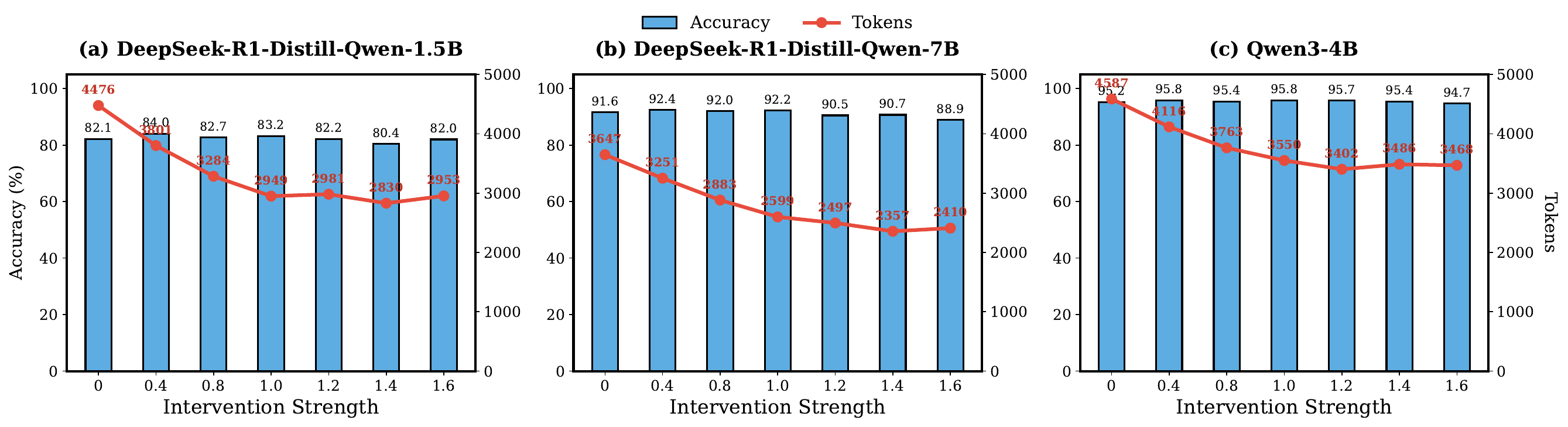} 
    \caption{The impact of intervention strength $\lambda$ on Pass@1 accuracy and average token count across three models on the MATH500 dataset.}
    \label{fig:ablation_studies_step}
\end{figure}

We evaluate the effect of the intervention strength $\lambda$, which controls the magnitude of the state correction. As shown in Figure \ref{fig:ablation_studies_step}, increasing $\lambda$ up to 1.0 yields  a substantial reduction in token consumption without compromising the accuracy. However, exceeding this threshold ($\lambda > 1.0$) introduces a clear trade-off: overly aggressive interventions provide only marginal token savings while degrading reasoning accuracy, likely due to latent state over-correction. Finally, $\lambda=1.0$ offers a favorable balance across different models, and thus we adopt it as our default setting.

\subsection{Impact of Steering Layer} 
\begin{figure}[htbp]
    \centering
    \includegraphics[width=\textwidth]{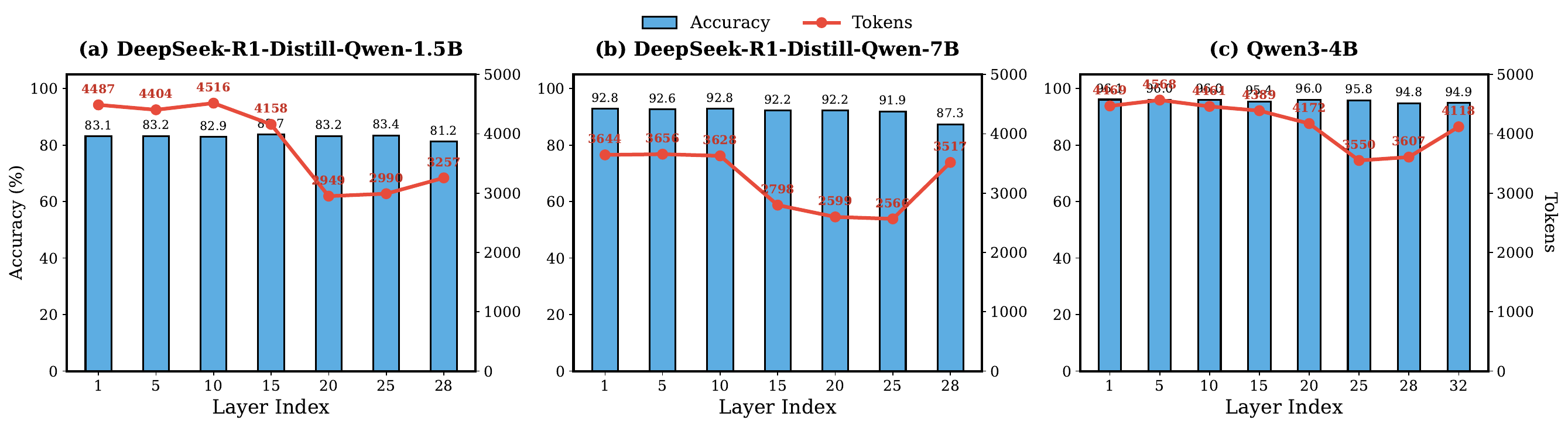} 
    \caption{The impact of the steering layer index on Pass@1 accuracy and average token count across three models on the MATH500 dataset.}
    \label{fig:ablation_studies_layer}
\end{figure}

Figure \ref{fig:ablation_studies_layer} illustrates the sensitivity of the performance to the chosen intervention layer. Intervening at middle-to-deep layers generally yields better performance, as these layers usually capture more abstract reasoning features \citep{liu2024fantastic}. Based on the empirical results, we set the steering layer to 20 for both the DeepSeek-R1-Distill-Qwen-1.5B and 7B models, and to 25 for the Qwen-4B model.

\subsection{Impact of PCA Dimension $k$}
\begin{figure}[htbp]
    \centering
    \includegraphics[width=\textwidth]{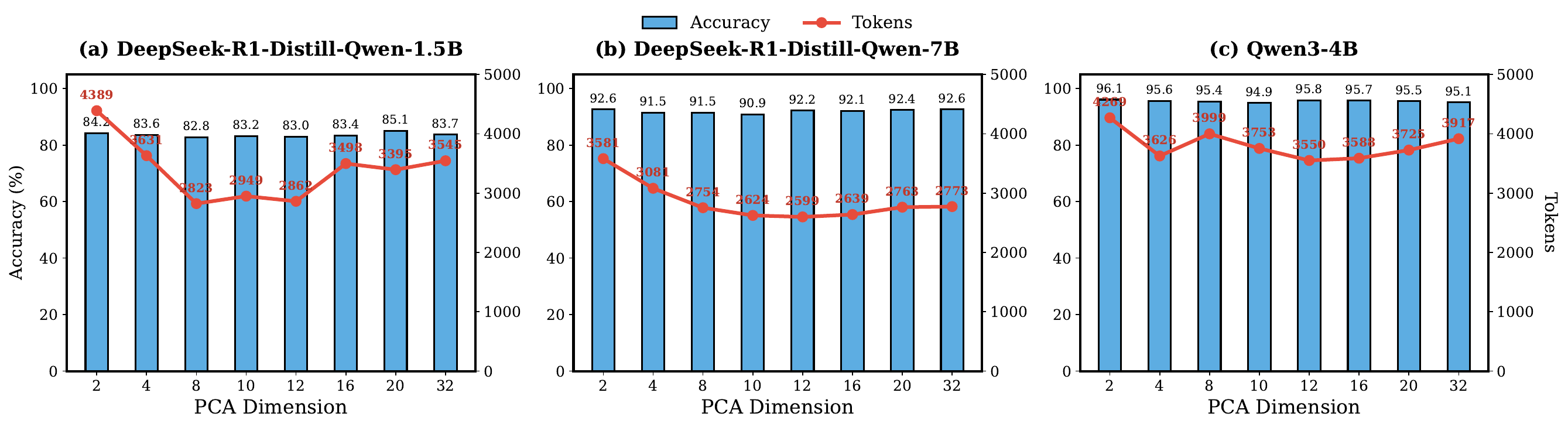} 
    \caption{The impact of the PCA dimension $k$ on Pass@1 accuracy and average token count across three models on the MATH500 dataset.}
    \label{fig:ablation_studies_dim}
\end{figure}

As shown in Figure \ref{fig:ablation_studies_dim}, we investigate the effect of the latent space dimension $k$ on the reasoning performance. A very low dimension may lead to severe information loss, making it difficult to construct a efficient reasoning set. Conversely, a high dimension introduces unnecessary noise and complicates the projection process without bringing notable performance gains. Since the performance remains relatively stable within the middle range, we select $k = 10$ for the DeepSeek-R1-Distill-Qwen-1.5B model, and $k = 12$ for both the DeepSeek-R1-Distill-Qwen-7B and Qwen3-4B models.

\subsection{Impact of Boundary Proportion $\alpha$} 
\begin{figure}[htbp]
    \centering
    \includegraphics[width=\textwidth]{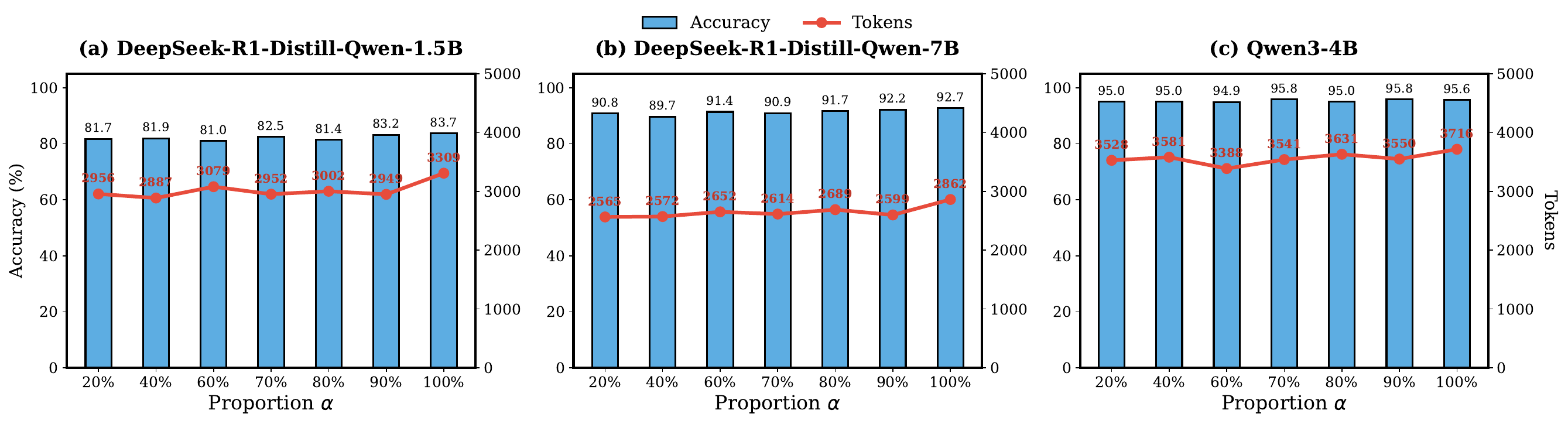} 
    \caption{The impact of the boundary shift proportion $\alpha$ on Pass@1 accuracy and average token count across three models on the MATH500 dataset.}
    \label{fig:ablation_studies_coverage}
\end{figure}

Figure \ref{fig:ablation_studies_coverage} demonstrates the influence of the boundary shift proportion $\alpha$. When $\alpha$ is small, the constructed set $\mathcal{H}$ is highly restricted. This leads to relatively aggressive projections that can inevitably discard useful semantic variations, resulting in a slightly weaker accuracy. On the other hand, when $\alpha = 100\%$, the token reduction becomes less pronounced. This is likely because a fully inclusive proxy set may also encompass some ``overthinking'' behaviors such as redundant double-checking, making the intervention conservative. Therefore, we select $\alpha = 90\%$ as a default setting across all models.

\subsection{Impact of Cluster Number $M$}
\begin{figure}[htbp]
    \centering
    \includegraphics[width=\textwidth]{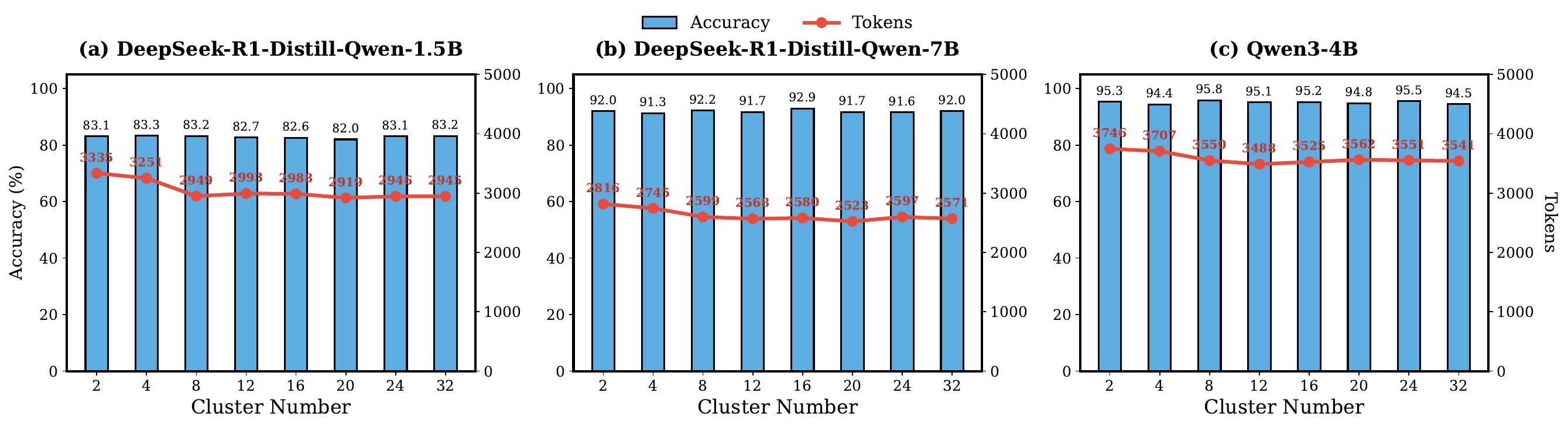} 
    \caption{The impact of the cluster number $M$ (number of linear constraints) on Pass@1 accuracy and average token count across three models on the MATH500 dataset}
    \label{fig:ablation_studies_cluster}
\end{figure}

Figure \ref{fig:ablation_studies_cluster} illustrates the effect of $M$, which defines the number of half-space constraints used to form the set $\mathcal{H}$. Empirically, a small $M$ yields limited efficiency gains, which may be due to the resultant set forming an overly loose boundary. As $M$ increases, the performance improves, likely because the boundary becomes tighter and more refined. However, a larger $M$ introduces additional linear constraints into the QP problem, inevitably increasing the computational latency during inference. To balance reasoning efficiency and computational overhead, we set $M = 8$ across all models.

\section{Limitations} \label{app:limit}
Several aspects of our study warrant future investigation. First, our current empirical evaluations are constrained by computational resources to reasoning models with up to 14B parameters. It remains an open question whether the geometric concentration of efficient reasoning steps observed in this work persists in much larger models (e.g., >70B parameters), which may exhibit more complex latent space. Second, while our method successfully improve the reasoning performance on general reasoning tasks such as mathematics and coding, its efficacy on highly specialized professional domains, such as complex legal argumentation or medical diagnosis, requires further evaluation.

\section{Broader Impacts}
\label{app:broader_impacts}
Our work focuses on achieving efficient reasoning and presents several potential positive societal impacts. Primarily, by reducing redundant token generation, our method drastically lowers the computational cost and energy consumption associated with deploying advanced AI systems. This efficiency gain can contribute to a reduced carbon footprint for AI operations. Furthermore, by alleviating the hardware requirements for high-performance reasoning, our approach makes powerful reasoning models more accessible to researchers, educators, and organizations with limited computational resources. As our research focuses on a generic optimization for inference efficiency, it serves as foundational research and does not introduce any direct paths to specific negative applications beyond the inherent capabilities of the underlying base models.
\vfill

\section{Case Study}

\begin{casebox}{Case 1}

\noindent
\textbf{Prompt:} Jamaar loves fresh fruit and is headed to the store with \$10 he earned mowing lawns. Including tax, peaches and pears are \$.5 each, apples are \$.75 each, kiwis are \$1, and plums cost \$.25 each. If he has already purchased 3 peaches, 4 pears, 2 kiwis, and 5 apples, how many plums can he buy? \hfill \textbf{Gold: 3}

\vspace{2mm}
\noindent\textcolor{gray!30}{\rule{\linewidth}{1pt}}
\vspace{2mm}

\noindent 
\begin{vanillabox}{Vanilla Output (673 Tokens)}
    \scriptsize 
    
    Alright, so I've got this problem here about Jamaar buying fruit with money he earned. Let me try to figure this out step by step. Sometimes math problems can be a bit overwhelming, but breaking them down usually helps.

    Let me list out the information given:
    \begin{itemize}[leftmargin=*, itemsep=0pt, parsep=0pt]
        \item \textbf{Total money Jamaar has:} \$10
        \item \textbf{Cost of each fruit:}
        Peaches: \$0.5 each; Pears: \$0.75 each; Kiwis: \$1 each; Plums: \$0.25 each
        \item \textbf{Fruits already purchased:}
        3 peaches, 4 pears, 2 kiwis, 5 apples
    \end{itemize}

    I need to calculate how much money he has spent on the fruits he already bought and then subtract that from his total to see how much he has left. After that, I can figure out how many plums he can buy...

    1. \textbf{Peaches:} He bought 3 peaches, each costing \$0.5. So, 3 $\times$ \$0.5 = \$1.50
    
    2. \textbf{Pears:} He bought 4 pears, each costing \$0.75. So, 4 $\times$ \$0.75 = \$3.00
    
    3. \textbf{Kiwis:} He bought 2 kiwis, each costing \$1. So, 2 $\times$ \$1 = \$2.00
    
    4. \textbf{Apples:} He bought 5 apples, each costing \$0.25. So, 5 $\times$ \$0.25 = \$1.25

    Now, let me add up all these amounts to find the total money spent so far. \\
    Total spent = \$1.50 + \$3.00 + \$2.00 + \$1.25
    
    First, \$1.50 + \$3.00 = \$4.50 \\
    Then, \$4.50 + \$2.00 = \$6.50 \\
    Next, \$6.50 + \$1.25 = \$7.75

    So, he's spent a total of \$7.75 on the fruits he already purchased.

    Now, let's subtract this from his total money: \\
    Money left = \$10.00 - \$7.75 = \$2.25

    Now, each plum costs \$0.25, so I need to find out how many plums he can buy with \$2.25. \\
    Number of plums = \$2.25 / \$0.25
    
    Let me think, \$0.25 goes into \$1.00 four times, right? So, \$0.25 goes into \$2.25 how many times? \\
    Well, 0.25 $\times$ 9 = 2.25 because 9 $\times$ 0.25 is 2.25. 
    
    Wait, let me double-check that. \\
    0.25 $\times$ 10 = 2.50, which is more than 2.25, so it must be 9. Yes, that makes sense. So, 9 plums.

    Wait a second, let me just make sure I didn't make a mistake in the initial calculations. \\
    Total spent on peaches: 3 $\times$ 0.5 = 1.50. Correct. \\
    Pears: 4 $\times$ 0.75 = 3.00. Correct. \\
    Kiwis: 2 $\times$ 1 = 2.00. Correct. \\
    Apples: 5 $\times$ 0.25 = 1.25. Correct.

    Total money: 10.00 - 7.75 = 2.25. Correct. \\
    Number of plums: 2.25 / 0.25 = 9. Correct.

    So, I think that's right. Therefore, Jamaar can buy 9 plums with the remaining money.
    \verb|</think>| \\

    Jamaar has \$10 and has already spent \$7.75 on peaches, pears, kiwis, and apples. With the remaining \$2.25, he can buy 9 plums. \\
    
    \textbf{Answer:} Jamaar can buy \boxed{9} plums.
\end{vanillabox}\hfill
\begin{stirbox}{Ours Output (221 Tokens)}
    \scriptsize 
    
    First, calculate the total amount of money Jamaar has spent on peaches, pears, kiwis, and apples.

    Peaches: 3 peaches $\times$ \$0.50 = \$1.50 \\
    Pears: 4 pears $\times$ \$0.50 = \$2.00 \\
    Kiwis: 2 kiwis $\times$ \$1.00 = \$2.00 \\
    Apples: 5 apples $\times$ \$0.75 = \$3.75 

    Add these amounts to find the total expenditure: \\
    \$1.50 + \$2.00 + \$2.00 + \$3.75 = \$9.25 

    Subtract this from the total money to find the remaining amount: \\
    \$10.00 - \$9.25 = \$0.75 

    Finally, divide the remaining money by the cost per plum to determine how many plums Jamaar can buy: \\
    \$0.75 $\div$ \$0.25 = 3 plums. \\
    \verb|</think>|

    \vspace{3mm}
    \textbf{**Solution:**}
    
    Jamaar has a total of \$10 to spend. He has already purchased various fruits, and we need to determine how many plums he can buy with the remaining money.

    \textbf{1. Calculate the total cost of the fruits already purchased:}
    \begin{itemize}[leftmargin=*, itemsep=0pt, parsep=0pt]
        \item \textbf{Peaches:} 3 $\times$ \$0.50 = \$1.50
        \item \textbf{Pears:} 4 $\times$ \$0.50 = \$2.00
        \item \textbf{Kiwis:} 2 $\times$ \$1.00 = \$2.00
        \item \textbf{Apples:} 5 $\times$ \$0.75 = \$3.75
    \end{itemize}

    \textbf{2. Sum up the costs:}
    \[ \$1.50 + \$2.00 + \$2.00 + \$3.75 = \$9.25 \]

    \textbf{3. Determine the remaining money:}
    \[ \$10.00 - \$9.25 = \$0.75 \]

    \textbf{4. Calculate the number of plums Jamaar can buy:} \\
    Each plum costs \$0.25. 
    \[ \frac{\$0.75}{\$0.25} = 3 \]

    \textbf{Final Answer:} 
    \[ \boxed{3} \]
\end{stirbox}

\end{casebox}



\end{document}